\documentclass[12pt]{article}

\usepackage{color}
\usepackage{amsmath,amssymb,amsthm}
\usepackage{graphicx,subfig}
\usepackage{grffile}
\usepackage[authoryear]{natbib}
\usepackage{algpseudocode}
\usepackage{algorithm}
\usepackage{algorithmicx}
\usepackage[colorlinks,citecolor=blue,urlcolor=blue]{hyperref}

\newtheorem{theorem}{Theorem}
\newtheorem{lemma}{Lemma}

\newtheorem{corollary}{Corollary}
\newtheorem{definition}{Definition}[section]

\newcommand{\EP}{\,\mathbb{P}}
\newcommand{\EE}{\,\mathbb{E}}
\newcommand{\Breg}{{\mathbf{D}}}
\newcommand{\breg}{{\mathbf{\Delta}}}

\newcommand{\bsb}{\boldsymbol}
\newcommand{\rd}{\,\mathrm{d}}

\newcommand{\bsbX}{{\boldsymbol{X}}}

\newcommand{\bsby}{{\boldsymbol{y}}}
\newcommand{\bsbY}{{\boldsymbol{Y}}}

\newcommand{\bsbI}{{\boldsymbol{I}}}

\newcommand{\bsbZ}{{\boldsymbol{Z}}}

\newcommand{\bsbSig}{{\boldsymbol{\Sigma}}}

\newcommand{\bsbXi}{{\boldsymbol{\Xi}}}

\newcommand{\bsbD}{{\boldsymbol{D}}}

\newcommand{\bsba}{{\boldsymbol{\alpha}}}
\newcommand{\bsbA}{{\boldsymbol{A}}}

\newcommand{\bsbC}{{\boldsymbol{C}}}

\newcommand{\bsbM}{{\boldsymbol{M}}}

\newcommand{\bsbE}{{\boldsymbol{E}}}

\newcommand{\bsbdelta}{{\boldsymbol{\delta}}}

\newcommand{\bsbS}{{\boldsymbol{S}}}

\newcommand{\bsbth}{{\boldsymbol{\theta}}}
\newcommand{\bsbTh}{{\boldsymbol{\Theta}}}

\newcommand{\bsbW}{{\boldsymbol{W}}}

\newcommand{\bsbB}{{\boldsymbol{B}}}

\newcommand{\bsbDelta}{{\boldsymbol{\Delta}}}

\newcommand{\real}{\mathbb{R}}

\DeclareMathOperator*{\argmin}{argmin}

\graphicspath{{figs/}{./}}
\DeclareGraphicsExtensions{.jpg,.pdf,.mps,.png,.eps}

\DeclareMathOperator*{\diag}{diag}
\DeclareMathOperator*{\sign}{sign}

\title{Indirect Gaussian Graph Learning \\beyond Gaussianity}
\author{Yiyuan She, Shao Tang, and Qiaoya Zhang\\Department of Statistics, Florida State University}
\date{}

\begin{document}

\maketitle

\begin{abstract}
This paper studies how to capture   dependency graph structures   from real data which may not be  multivariate Gaussian. Starting from marginal loss functions  not necessarily derived from   probability distributions, we utilize an additive over-parametrization with shrinkage to   incorporate     variable dependencies into the criterion. An  iterative Gaussian graph learning algorithm is proposed with   ease in implementation. Statistical analysis shows that  the estimators achieve satisfactory accuracy  with the error measured in terms of a proper Bregman divergence. Real-life examples  in different settings are given   to demonstrate the efficacy of the proposed methodology.
\end{abstract}

\section{Introduction}\label{sec:intro}

Given multivariate observations for a number of random variables, learning      dependency graph structures has widespread   applications in machine learning, bioinformatics, and social studies.
This gives rise to the undirected Gaussian Graphical Learning (GGL, for short), where the existence of an edge corresponds to a nonzero entry in the inverse covariance matrix. Much effort has been devoted to \textit{sparse} inverse covariance estimation, where sparsity can be achieved by enforcing an $\ell_1$-norm penalty on the (off-diagonal) entries of the inverse covariance.
\cite{meinshausen2006high} approached the problem by neighborhood selection. Their method results in an estimate that is not necessarily symmetric or positive-definite despite its speed from the simple design. Later methods mainly focus on solving the regularized GGL criterion using a variety of optimization techniques. See, for example, \cite{yuan2007model,banerjee2008model,friedman2008sparse,witten2011new,mazumder2012exact,oztoprak2012newton,hsieh2014sparse} among many others.

Despite the popularity and accessibility of GGL algorithms, the multivariate Gaussianity assumption is a rather stringent one, and becomes inappropriate when the data are fat-tailed, skewed, or discrete. Graph learning   becomes far more challenging when the observed data are not   Gaussian.
There are   a series of semiparametric papers that use  copula transformations      \citep{liu2009nonparanormal,dobra2011copula,liu2012nonparanormal,xue2012regularized}. Merely using   data ranks could be lossy and may show limited robustness against gross outliers. Recently, \cite{Fan2017b} proposed a latent Gaussian copula model to treat mixed binary and continuous data. The bridge function that links Kendall's $\tau$ to latent correlations varies from case to case, and the   additional cutoff parameters  are  hard to estimate in the presence of   mean parameters. Overall, these studies do not cater to discrete data  (often multi-leveled)   very well. The class of score matching methods \citep{hyvarinen2005estimation,lin2016,Yu2016} suffers the same issue.

Many parametric methods build upon  Markov random fields (MRFs).
The most popular and well-studied model for binary data is probably the Ising model \citep{ising1925beitrag}. However, it is much more difficult to optimize the  penalized likelihood compared to its Gaussian counterpart, because the so-called normalizing constant, which is a function of interaction coefficients, can be computationally  intractable, and so people resort to various approximations. In  light of \cite{meinshausen2006high}, \cite{ravikumar2010high} turned to  node-wise $\ell_1$-penalized logistic likelihoods, but the method shares the same pitfalls and requires post-processing. 
A group of pseudo (or composite) likelihood based methods \citep{hofling2009estimation,xue2012nonconcave}, developed upon \cite{besag1975statistical}, use the sum  of conditional likelihoods in place of the genuine joint likelihood function.

Compared to   binary data,   Poisson data---prevalent in text, genomic sequencing, site-visit, and climate problems---has been a much harder category to model conditional dependency upon, thus enjoys less success. Combinatorial approaches \citep{madigan1995bayesian,lauritzen1996graphical}
 have been proposed, but they become easily intractable for even a moderate number of variables.  \cite{karlis2003algorithm} modeled  observed counts as partial sums of a series of   Poisson random variables, 
and thus {only} positive correlations can be taken into account.   On the contrary,    \cite{yang2012graphical, yang2015graphical}, also developed upon \cite{besag1975statistical}, can only pick negative conditional dependencies.    To address the issues,   some efforts have been taken to truncate large counts   or  alter the distribution function, see, e.g.,  \cite{yang2013poisson}. 
Some of these modifications    seem       ad-hoc and may fail to produce a  joint likelihood with applicability.  Recently, some studies have been performed  to deal with mixed types of data, where not all variables follow the same type of distribution \citep{yang2014mixed,Leehastie14}.

There exist yet other works targeting on non-Gaussian graph learning.   In reality, these models,   starting from a joint distribution or conditional likelihoods  defined for a large number of variables,   may not hold exactly, due to  data imperfections (e.g., heavy tails and skewness). Even without such issues, we have seen that the Poisson MRF has  severe  limitations
in  modeling  dependencies on count data. On the other hand, given each individual variable, practitioners often have a clear idea of what makes a proper discrepancy measure.
Our launching point is   the pre-specified \textit{marginal} loss functions. It is worth noting that these losses may not belong to the exponential family or not even correspond to  any  likelihoods. So the crucial problem here  is how  to combine the given losses in a smart fashion, rather than  figuring out a   multivariate  distribution for a large number of nodes.

   Not so surprisingly,   because learning the  {precise} conditional dependence structure is challenging even for binary data or Poisson data, approximations have to be  made. We will indeed make   simplifying assumptions and study an   easier problem, which, in turn,    gains  \textbf{flexibility} and  \textbf{implementation ease}. One novelty of our work is to wrap up    high-order statistics  into the mean using  a technique of additive over-parametrization with shrinkage, which corresponds to a random effects model in the Gaussian setting. The proposed  graph learning   framework is able to accommodate various data types, and calls   Gaussian graph learning  iteratively  to conquer   non-quadratic losses.

In the rest of the paper,  Section \ref{sec:method} describes in detail the indirect methodology of how to take  associations into account, and argues its equivalence to the   weighted fashion in GGL. Section \ref{sec:algorithm} develops an iterative  GGL   algorithm and introduces a convenient trick to handle     Poisson data. Section \ref{sec:theory} performs some nonasymptotic studies.
Section \ref{sec:exp} demonstrates the performance of the proposed method with some  real-life data examples.
We conclude in Section \ref{sec:conc}.

In the rest of the paper, the following notation and symbols will be  used. Given a matrix $\bsbA  \in \mathbb R^{n\times m}$,   $\|\bsbA\|_F$ and $\|\bsbA\|_2$  denote its Frobenius norm and spectral norm, respectively. We   denote its elementwise    $\ell_1$-norm    by $\|\bsbA\|_{1} = \| \mbox{vec} (\bsbA)  \|_1 $, where $\mbox{vec}$ is the standard vectorization operation.   Finally, for any positive semi-definite $\bsbA$, $\bsbA^{1/2}$ denotes its (matrix) square root.

\section{Generalized Indirect Dependency Learning}\label{sec:method}

Given an observation matrix $\bsbY\in\mathbb R^{n\times m}$ with $n$ observations of $m$ variables, and an estimate $\bsbTh$ of the same dimensions (i.e., an $n\times m$ matrix), under the independence assumption, it is natural to describe the overall discrepancy    by
\begin{align}
\bar{l}(\bsbTh; {\bsbY}) = \sum_{k=1}^m l_k (\bsbth_k, \bsby_k), \label{lbardef}
\end{align}
where  $l_k$ is a pre-specified loss for the $k$th variable. A special case is
\begin{align}
\bar{l}(\bsbTh; {\bsbY}) = \sum_{i,k} l(\theta_{i,k},y_{i,k}). \label{simpleunivloss}
\end{align}
Such loss functions  can be customized by users depending on the problem  of interest. They are not necessarily identical in some applications. 
 $\bsbTh$   represents the systematic component, e.g., $\bsbTh = \bsbX\bsbB$ with $\bsbX\in \mathbb R^{n\times p}$  the design matrix and $\bsbB$ the corresponding coefficient matrix, but may not be on the same scale of $\bsbY$. For example, in Generalized Linear Models (GLMs) \citep{Agresti2012}, $\bsbTh = g(\EE(\bsbY))$, where $g(\cdot)$ is a link function.

The $m$ random variables   represented by the columns of $\bsbY$ are however often dependent and  it could be very  hard to build an MRF. (Recall that $l_k$ may not be associated with any distribution.) With no permission to modify any marginal loss, we aspire to   incorporate dependencies into the criterion and learn   a meaningful sparse association graph.

\subsection{Additive over-parameterization with shrinkage}
Let's motivate our framework in the multivariate Gaussian scenario. Assume $\mathrm{vec}(\bsbY) \sim \mathcal{N}(\mathrm{vec}(\bsbM), \bsb\Sigma \otimes \bsbI)$, where $\bsb\Sigma \in \mathcal{S}_{++}^m$---the set of all positive definite matrices of size $m\times m$, and $l(\theta, y) = (\theta - y)^2/2$.
In this model, the rows of $\bsbY$ (samples) are independent while the column dependency is characterized by $\bsbSig$, or $\bsbW := \bsbSig^{-1}$.   It is well known that  $w_{ij}=0$ in the inverse covariance matrix (or precision matrix)  indicates that  the $i^{th}$ and the $j^{th}$ variables are independent given all other variables.

Suppose the mean  $\bsbM$ is given or can be well estimated.
In order to estimate $\bsbW$, one can solve
\begin{equation}\label{ggl_full}
\begin{aligned}
&\min_{\bsbW \in \mathcal{S}_{++}^m} \frac{1}{2} Tr\{ (\bsbY-\bsbM)\bsbW (\bsbY-\bsbM)^T\} - \frac{n}{2}\log \det \bsbW +   P_W(\bsbW; \lambda_W),
\end{aligned}
\end{equation}
or the following form after introducing $\bsbS = (\bsbY-\bsbM)^T(\bsbY-\bsbM)/n$ (which   is the sample covariance when $ \bsbM = \bsb1  \bsb1^T \bsbY/n $)
\begin{equation}
\begin{aligned}
\min_{\bsbW \in \mathcal{S}_{++}^m} Tr\{\bsbS \bsbW\} - \log \det \bsbW + 2 P_W(\bsbW; \lambda_W)/n,  \label{ggl}
\end{aligned}
\end{equation}
where    $P_W$  is a sparsity-inducing penalty.     A popular choice for the penalty is the $\ell_1$-norm function, the resulting  problem  termed the graphical lasso \citep{friedman2008sparse}.

The quadratic loss in \eqref{ggl_full} utilizes a weighting scheme to capture associations. The technique applies more generally in MRFs with $w_{ij}$ denoting interaction coefficients.  For an arbitrary loss  one   could  write $l(\theta_{ij}, y_{ij}) = [l(\theta_{ij}, y_{ij})]^{1/2}  [l(\theta_{ij}, y_{ij})]^{1/2}$ to mimic the quadratic form. But it  does not have sound theoretical support, nor does it lead to simple computation.
Instead, we propose to modify the \underline{mean} by adding a shift term $\bsbC ( \bsbI - \phi\bsbW)^{1/2}$, and so $\bsbM$ is replaced by
\begin{align}
\bsbTh=\bsbM + \bsbC(  \bsbI - \phi \bsbW)^{1/2}.
\end{align}
Here, $\bsbC \in \mathbb{R}^{n \times m}$ is an unknown component, and $\phi$ takes a small enough positive value such that $ \bsbI - \phi\bsbW$ is positive semi-definite. Of course, the additive representation is \textit{over-complete} since $\bsbC$ is already of the same size of $\bsbY$. We append an $\ell_2$-type penalty  $Tr\{\bsbC \bsbW \bsbC^T\}/2$  to the loss, and define the additive over-parametrization with shrinkage (\textbf{AOS}) criterion    in  $\bsbW$ and $\bsbC$ jointly
\begin{gather} \label{ggl_aos}
\begin{split}
        \min_{0\preceq\bsbW \preceq \phi^{-1}\bsbI, \, \bsbC } \ \
        &\frac{1}{2 \phi} \|\bsbY  - \bsbM  - \bsbC ( \bsbI - \phi  \bsbW)^{1/2}\|_F^2
        +\frac{1}{2  } Tr\{\bsbC \bsbW \bsbC^T\}
        \\
        &-\frac{n}{2}\log \det \bsbW  + P_W(\bsbW ; \lambda_W).
\end{split}
\end{gather}
Interestingly, \eqref{ggl_aos} is an equivalent formulation to \eqref{ggl_full}.
Theorem \ref{lemma:equiv} shows a general result when jointly estimating $\bsbM$ and $\bsbW$.
Let  $\mathcal{Z}=  \mathcal{S}_{++}^m(\phi)\times \real^{n \times m} $ with $\mathcal{S}_{++}^m(\phi) = \{\bsbW\in\mathcal{S}_{++}^m: \bsbW   \preceq   \bsbI/\phi\}$.

\begin{theorem} \label{lemma:equiv}
        Let $\bsbY \in \mathbb{R}^{n \times m}$,
        $\bsbC \in \mathbb{R}^{n \times m}$,
        $\bsbM \in \mathcal{X} \subset \real^{n \times m}$.   Suppose the solutions to problem \eqref{f2}
          are  (uniformly) bounded. Then,  as long as    $\phi$    is set small enough,    the optimization problem
        \begin{equation}\label{f1}
        \begin{aligned}
        \min_{\bsbM \in \mathcal X, ( \bsbW, \bsbC) \in \mathcal{Z} } \
        &  \frac{1}{2 \phi} \|\bsbY  - \bsbM - \bsbC ( \bsbI - \phi \bsbW)^{1/2}\|_F^2
        +\frac{1}{2 } Tr\{\bsbC \bsbW \bsbC^T\}
        \\
        &-\frac{n}{2}\log \det \bsbW + P_M(\bsbM; \lambda_M) + P_W(\bsbW; \lambda_W)
        \end{aligned}
        \end{equation}
        is equivalent to
        \begin{equation}\label{f2}
        \begin{aligned}
        \min_{\bsbM\in \mathcal X,   \bsbW \in \mathcal{S}_{++}^m}  & \frac{1}{2 } Tr\{(\bsbY  - \bsbM) \bsbW(\bsbY  - \bsbM)^T\} - \frac{n}{2}\log \det \bsbW
        \\& + P_M(\bsbM; \lambda_M) + P_W(\bsbW; \lambda_W)
        \end{aligned}
        \end{equation}
        in the sense that the optimal solutions $(\bsbM, \bsbW)$ are the same.
\end{theorem}

From a Bayesian perspective,  the auxiliary matrix     $\bsbC$  can be viewed as random effects,  with a proper right-design matrix $( \bsbI - \phi \bsbW)^{1/2}$ to introduce between-column dependencies, i.e., $
 \mbox{vec} (\bsbY) | \bsbC \sim \mathcal N(\mbox{vec}( \bsbM +\bsbC(\bsbI - \phi\bsbW)^{1/2}),   \phi \bsbI\otimes \bsbI),     \mbox{vec}(\bsbC)  \sim \mathcal N(\bsb{0},  \bsbW^{-1}\otimes \bsbI).$  Then the conclusion  is  perhaps easier to understand  due to  the following identity
$$
\bsbW^{-1} = (\bsbI - \phi \bsbW)^{1/2} \bsbW^{-1} (\bsbI - \phi \bsbW)^{1/2} + \phi \bsbI.
$$But     in general, integrating $\bsbC$ out  is   a formidable task even when $m$ is moderate. The equivalence of \eqref{f1} and \eqref{f2} is  actually built by solving a   ridge-type {\textbf{optimization}} problem; see the proof for detail.
As opposed to the weighting mechanism, our \textit{indirect} way, by introducing an auxiliary matrix, enables  characterization of dependencies {without} the need of modifying the    loss.

 Back to the general case, we propose the  following  criterion
\begin{equation}
\begin{aligned}
\min_{ \bsbM\in \mathcal X, (\bsbW, \bsbC) \in \mathcal{Z}} \
 &   {\phi}^{-1} \, \bar l(  \bsbM + \bsbC(  \bsbI - \phi\bsbW)^{1/2};  \bsbY)
+\frac{1}{2  } Tr\{\bsbC\bsbW \bsbC^T\}\\&-\frac{n}{2}\log \det \bsbW +  P_M(\bsbM; \lambda_M) + P_W(\bsbW; \lambda_W).
\label{crit:raw}
\end{aligned}
\end{equation}
The learning framework allows for customizing   marginal losses (recall $\bar{l}  = \sum  l_k  $),  which is    helpful in handling mixed types of data.

The mean term  $\bsbM$  varies in different scenarios: $\bsbM$ is often  $\bsbX\bsbB$ in the presence of a design matrix,     and when there are no predictors, $\bsbM = \bsb1 \bsba^T$. In the rest of the paper, we always assume that $\bsbM$ is known (or can be well estimated beforehand) unless otherwise stated, and   focus on the estimation of $\bsbW$, the problem referred to as the \textit{indirect} Gaussian Graph Learning (\textbf{iGGL}) (the prefix `i' also stands for \emph{iterative} in algorithm development).
Experience shows that  $\bsbW$    provides a useful instrument for capturing some interesting association structures in practice.

In mixed   graph learning where   $l_k$ are  not all the same, it might be helpful to perform a scale calibration. Concretely, under the assumption that  $\bsbM = \bsb1 \bsba^T$   with $\bsba = [\alpha_1, \ldots, \alpha_m]^T$,   $l_k(\bsbth_k) = l_k(\theta_{1, k}) + \cdots + l_k(\theta_{n, k})  $, and $l_k''(\alpha_k)$ exist, we could  scale $l_k(\cdot)$ by $l_k''(\alpha_k)$. The intuition comes from the following theorem (which  can be easily adapted to  show a large-$n$  asymptotic result). Define   $\breg(\bsbTh; \bsbM)=[\bsbdelta_1, \ldots, \bsbdelta_m]$ with $
\bsbdelta_k= \nabla l_k(\bsbth_k) - \nabla l_k(\bsb1 \alpha_k) -    l_k''(\alpha_k)(\bsbth_k - \bsb1 \alpha_k)
$,   the remainder when expanding $\nabla\bar l$ to the second order.

\begin{theorem} \label{lemma:asym}
Consider the problem of       $\min_{ (\bsbW, \bsbC) \in \mathcal{Z}}{\phi}^{-1} \, \bar l(  \bsbM + \bsbC(  \bsbI - \phi\bsbW)^{1/2};  \bsbY)\allowbreak
+\frac{1}{2  } Tr\{\bsbC\bsbW \bsbC^T\} -({n}/{2})\log \det \bsbW$.
Define \begin{align}
  \bsbSig_n( \bsbM, \breg) =   \frac{(\nabla \bar l (\bsbM)+ \breg)^T  (\nabla \bar l (\bsbM)+ \breg) }{n}.\label{Sigmadef}\end{align}
Then there exists a stationary point $(\hat \bsbW, \hat \bsbC)$ such that  $\hat \bsbSig_n := \bsbSig_n( \bsbM, \breg(\hat\bsbTh; \bsbM))$ can be represented in terms of $\hat \bsbW$:  \begin{align}
 \hat\bsbSig_n = &[\bsbD + \phi \hat \bsbW (\bsbI - \bsbD)]^{}   \hat \bsbW^{-1} [\bsbD + \phi \hat \bsbW (\bsbI - \bsbD)]^{},\label{esteq1}
 \end{align}
where    $\bsbD=\diag\{l_k''(\alpha_k)\}$.
 In particular, if $\bsbD = d \bsbI$,
\begin{align}
 \hat\bsbSig_n =     \hat \bsbW^{-1} [d \bsbI + (1-d) \phi \hat \bsbW  ]^{2}.\label{esteq1_special}
 \end{align}
\end{theorem}

Equations \eqref{esteq1} and  \eqref{esteq1_special}    provide some insight into inverse covariance recovery when only marginal losses $l_k$ of the joint model are available.
It is easy to see that $-\nabla \bar l(\bsbM)$ in \eqref{Sigmadef}
 gives the noise component; in fact, in the GLM case of  $\bar l(\bsbTh) = -\langle \bsbY, \bsbTh\rangle + \langle \bsb1, b(\bsbTh)\rangle$ (cf.  Section \ref{subsec:exs}), $$-\nabla \bar l(\bsbM) = \bsbY - b'(\bsbM) = \bsbY - \EE[\bsbY].$$    For Gaussian random variables, the higher-order remainder term $\breg$ is zero. When $n\rightarrow \infty$, $\bsbDelta$  can   be well controlled   under some   regularity conditions on   $l_k$, and so when    asymptotic normality holds, $\hat \bsbSig_n$ does resemble  the covariance matrix as expected.    In finite samples with    non-quadratic losses being applied,  $\hat \bsbSig_n$     necessarily contains higher-order terms (which is desirable).
According to   \eqref{esteq1_special},  when  $(1-d)\phi=0$ or  $d=1$,  $\hat \bsbW   =   \hat \bsbSig_n^{-1}$.
On the other hand,   as long as  $\phi$ takes a sufficiently small value,    $\hat \bsbW$ is approximately $\bsbD  \hat\bsbSig_n^{-1} \bsbD $ from \eqref{esteq1}, i.e., $\hat\bsbSig_n^{-1}$ up to some row/column scalings. 

\subsection{Examples of loss functions }\label{subsec:exs}

The iGGL framework automatically incorporates dependency and  is universal in that it does not limit to a specific loss. In fact, one can safely use any marginal loss(es) as if the variables were independent. We give some examples to illustrate its applicability.

\paragraph*{Exponential family.}
Given a canonical GLM with link $g$, the (univariate) loss is given by $l(\theta, y) = -y\theta + b(\theta)$, where $b(\cdot)$ serves as the cumulant function and $b'(\cdot) = g^{-1}(\cdot)$. Then $\bar l  = - \langle \bsbY,   \bsbTh\rangle  +   \langle \bsb1,  b(\bsbTh) \rangle$, where  $b$ is applied componentwise.
Many distributions of interest in the family, e.g., Gaussian, Bernoulli, binomial and multinomial, have a cumulant function with bounded curvature: $b''(\cdot) \leq L$ for some constant $L$.
This property greatly simplifies computation, as will be demonstrated in Section \ref{ssec:surrogate}. 

\paragraph*{Robust losses.}
Another important class of loss functions beyond Gaussianity take robustness as a major concern.  Many popular alternatives to the quadratic loss are  defined via a  $\psi$-function:  $ l(\theta, y) = \int_0^{|\theta-y|} \psi(t) \rd t $ \citep{huber2011robust,hampel2011robust}.
Huber's $\psi$ is given by
\begin{align}
\psi(t) =
\begin{cases}
t, & |t| \leq c\\
c  \sign(t), & |t| > c,
\end{cases}
\end{align}
where $c=1.345 \sigma$ is recommended and    $\sigma$ is   some  robust estimate of the   standard deviation of errors.
 Tukey's bisquare is
\begin{align}
\psi(t) =
\begin{cases}
t\big[ 1-({\frac{t}{c}})^2\big]^2,  & |t| \leq c\\
0,                                                                           & |t| > c,
\end{cases}
\end{align}
where $c = 4.685 \sigma$ is recommended.
 Hampel's three-part $\psi$ is
\begin{align}
\psi(t) =
\begin{cases}
t,                        & |t| \leq a\\
a  \sign(t)      & a< |t| \leq b\\
a  \sign(t)\frac{c-|t|}{c-b}      & b< |t| \leq c\\
0       & |\theta - y| > c.
\end{cases}
\end{align}
where $a/(c-b)$ is often   at $1/2$.
It is worth mentioning that to bound the influence of outliers, both Tukey's $\psi$  and Hampel's $\psi$  are designed to be redescending, meaning that the associated loss functions are non-convex, which can be just as well accommodated by iGGL.

\paragraph*{Some classification losses.}
  The {Huberized hinge loss}  \citep{rosset2007piecewise} can be used in support vector machines to  reduce the misclassification error
\begin{align}
{l}(\theta, y) =
\begin{cases}
1 -  {c}/{2}- y \theta ,  & y\theta \leq 1-c\\
(1-y \theta)^2/(2c),                 & 1-c < y\theta \leq 1 \\
0,                                & y \theta >1,
\end{cases}\label{HuberHinge}
\end{align}
where  $y=\pm 1$ and $c$  is a parameter often taking value  $1$.
    Savage loss, ${l}(\theta, y) =
  (1+e^{2y\theta})^{-2}
$,  is widely used in boosting   \citep{Hamed2009}
and its gradient has Lipschitz constant $0.62$. Another interesting    Lorenz loss \citep{Barbu2016} takes the form of
\begin{align}
{l}(\theta, y) =
\begin{cases}
\log(1+(y\theta-1)^2),  & y\theta \leq 1\\
0,                      & y\theta >1.
\end{cases}
\end{align}
 Savage loss and Lorenz loss are more resistant to mislabeled samples than \eqref{HuberHinge}.
Clearly, these losses are not associated with distributions.

\section{Iterative GGL for Computation }\label{sec:algorithm}

Before    describing   the algorithm design in thorough detail,  it may
help the reader to check  the  pseudocode of the iterative Gaussian graph learning (iGGL) in Algorithm \ref{Alg:W}. The key step (line 5)   solves      an ordinary GGL problem,   after forming     $\bsbXi$ and $\bsbS$   on the basis of $\bsbTh$. Nicely, we will see that in the Gaussian case,   iGGL degenerates to GGL (and  so converges in one iteration).

\begin{algorithm}[t!]
        \caption{The iterative GGL (iGGL) algorithm}\label{Alg:W}
        \textbf{Input}:
        $\bsbY \in \real^{n \times m}$,  $\bsbM \in \real^{n \times m}$, $\lambda$,   $\nabla \bar l$ satisfying Lip-1;     $\bsbXi^{[0]}\in\mathbb R^{n\times m}$ (e.g.,   $\bsbXi^{[0]} = \bsbY$), $\bsbW^{[0]} \in \mathcal{S}_{++}^m$ (e.g., $  \mbox{diag}\{1/var(\bsby_k)\}$),  $\phi$ small enough (say $c  /\| \bsbW^{[0]}\|_2$ with $c=$\mbox{1e-3})
        \begin{algorithmic}[1]
                \State
                $\bsbTh^{[0]} \gets \bsbXi^{[0]} +\phi (\bsbM-\bsbXi^{[0]})\bsbW^{[0]} $;
                \While{  not converged }
                \State $k\gets k+1$;
                \State  $\bsbXi^{[k]} \gets \bsbTh^{[k-1]} - {\nabla \bar l(\bsbTh^{[k-1]})}$, $\bsbS^{[k]}\gets (\bsbXi^{[k]}-\bsbM)^T(\bsbXi^{[k]}-\bsbM)/ n $;
                \State ${\bsbW}^{[k]} \gets  \argmin_{\bsbW \in \mathcal{S}_{++}^m} -\log \det(\bsbW)+ Tr\{\bsbS^{[k]}\bsbW \} +   2 P(\bsbW; \lambda)/n$;
                \State  ${\bsbTh}^{[k]}\gets \bsbXi^{[k]} + \phi(\bsbM - \bsbXi^{[k]}) \bsbW^{[k]} $;
                \EndWhile \\
                \Return $\bsbW^{[k]}$.
        \end{algorithmic}
\end{algorithm}

\subsection{Linearization through $\bsbTh$ } \label{ssec:surrogate}

For simplicity, we assume that the gradient of each loss function is Lipschitz continuous:
\begin{align}
\|\nabla l_k (\bsbth_1) - \nabla l_k (\bsbth_2)\|_2 \leq L \|\bsbth_1 -\bsbth_2\|_2, \quad  \forall \bsbth_1, \bsbth_2 \label{lipcond}
\end{align}
where $L$ is a constant. Recall the optimization problem with $\bsbM$ given
\begin{equation}
\begin{aligned}
\min_{( \bsbW, \bsbC) \in \mathcal{Z}}
F:= &  {\phi}^{-1} \, \bar l( \bsbM + \bsbC(  \bsbI - \phi\bsbW)^{1/2} ;  \bsbY)
+\frac{1}{2  } Tr\{\bsbC\bsbW \bsbC^T\}\\ &-\frac{n}{2}\log \det \bsbW   + P(\bsbW ; \lambda),
\label{crit:raw}
\end{aligned}
\end{equation}
where $\phi$ is sufficiently small assumed throughout the section,  and $\bar l=\sum l_k$   as in  \eqref{lbardef}.

It might look straightforward to apply BCD to solve for $ \bsbW$ and $\bsbC$ alternatively. We take however  a different but efficient route to convert  the problem to GGL based on Theorem \ref{lemma:equiv}, where the key is to   linearize the first term in \eqref{crit:raw} through $\bsb\Theta = \bsbM + \bsbC(  \bsbI -\phi \bsbW)^{1/2}$ as a whole.
 Given the $k$th iterate $( \bsbW^{[k]},\bsbC^{[k]})$ and  $\bsbTh^{[k]}=\bsbM + \bsbC^{[k]}(  \bsbI - \phi \bsbW^{[k]})^{1/2}$,   construct a
surrogate function
\begin{align*}
& g(\bsbW, \bsbC ;  \bsbW^{[k]}, \bsbC^{[k]})
=      {\phi}^{-1}\,  \bar{l}( \bsbTh^{[k]} ; \bsbY) + {\phi}^{-1}  \langle \nabla_{\bsbTh} \bar{l}(\bsbTh^{[k]}), \bsbTh - \bsbTh^{[k]}\rangle \\& +  \frac{\rho}{2\phi}  \| \bsbTh - \bsbTh^{[k]}\||_F^2  + \frac{1}{2} Tr\{\bsbC\bsbW \bsbC^T\} -\frac{n}{2}\log \det \bsbW  + P(\bsbW; \lambda),
\end{align*}
where $ \nabla_{\bsbTh} \bar{l}$ is the gradient of $\bar{l}$ with respect to $\bsbTh$. In the uniform  case of \eqref{simpleunivloss},    $ \nabla_{\bsbTh} \bar{l}(\bsbTh;\bsbY) = [l'(\theta_{ij};y_{ij})]$.
Now define the $(k+1)^{th}$ iterate  as
\begin{align}
(\bsbW^{[k+1]}, \bsbC^{[k+1]})
= \argmin_{( \bsbW, \bsbC) \in \mathcal Z}
g( \bsbW, \bsbC;  \bsbW^{[k]}, \bsbC^{[k]}). \label{T+1}
\end{align}


\begin{theorem}
        \label{lemma:lip}
Assume the Lipschitz-gradient   condition   \eqref{lipcond}.
 Then,  as long as   $\rho \geq L$,   the sequence of iterates defined by \eqref{T+1} satisfies
$$
F( \bsbW^{[k+1]}, \bsbC^{[k+1]}) \le  F( \bsbW^{[k]}, \bsbC^{[k]}).
$$
That is, the objective function values
are non-increasing during the iteration.
\end{theorem}

See Appendix \ref{apdx:lip} for its proof which holds as long as $\nabla \bar l$ is Lipschitz.
The problem boils down to the $g$-optimization in  \eqref{T+1}, which  is  way simpler than  direct minimizing $F$. We rewrite the problem in the form of
\begin{equation}
\begin{aligned}
\min_{(\bsbW, \bsbC) \in \mathcal Z}  \
&  \frac{\rho}{2 \phi} \| \bsbM + \bsbC(  \bsbI -\phi \bsbW)^{1/2}  - \bsbXi^{[k+1]}\|_F^2
+\frac{1}{2 } Tr\{\bsbC \bsbW \bsbC^T\} \\
&-\frac{n}{2  }\log \det (\bsbW) +  {P(\bsbW; \lambda)},
\end{aligned} \label{func:opt}
\end{equation}
where
\begin{align}
\bsbXi^{[k+1]}=\bsbTh^{[k]} - \nabla_{\bsbTh} \bar{l}(\bsbTh^{[k]})/\rho \label{Xik}
\end{align}
and $1/\rho$ amounts to the step size.

The problem can be further simplified---under    condition \eqref{lipcond}, we can scale each loss function by $L$ beforehand and set  $\rho=1$. (The problem is much harder for $\rho>1$.)
For example, the Bernoulli deviance $l$ satisfies  $L = 1/4$, and we can use $4 l $ as  the input loss function.  Of course, when   $L \leq 1$, one does not have to perform the scaling to take $\rho = 1$, but this  will result in some   sacrifice in convergence speed.
The  quadratic loss has  $L= 1$, and the Lorentz loss satisfies $L=2$.

While the Lipschitz continuity on the gradient is desirable to achieve a universal step size, it is not absolutely necessary in implementation. One could apply some line search   with $F( \bsbW^{[k+1]}, \bsbC^{[k+1]}) \le g( \bsbW^{[k+1]},\allowbreak \bsbC^{[k+1]}; \allowbreak \bsbW^{[k]}, \bsbC^{[k]} )$ to get proper stepsizes to guarantee non-increasing objective function values.

\subsection{$\bsbW$-optimization} \label{ssec:W}

It remains to solve \eqref{func:opt} with $\rho=1$. Nicely, applying Theorem \ref{lemma:equiv} again (or plugging in   $\bsbC= ({\bsbXi}^{[k+1]} -\bsbM)( \bsbI - \phi\bsbW)^{1/2} $),   we are back to the GGL problem in computation
\begin{align}
 \min_{\bsbW\in \mathcal S_{++}^m}
&
\frac{n}{2 }
Tr\{ \bsbS^{[k+1]}
\bsbW  \}
-\frac{n}{2}\log \det (\bsbW) +  P(\bsbW; \lambda), \label{W_general_crit}
\end{align}
where $\bsbS^{[k+1]} =   (\bsbXi^{[k+1]} - \bsbM)^T(\bsbXi^{[k+1]} - \bsbM)/n$.

There is a rich collection of GGL algorithms in the literature. The $\ell_1$-penalized form of \eqref{W_general_crit} gives the   convex  {graphical lasso} problem.
   \cite{friedman2008sparse} partition the inverse covariance matrix and estimate its corresponding sparse rows/columns in a block-wise fashion. There are also some fast  second-order methods, see, e.g., 
   \cite{hsieh2014sparse},
\cite{oztoprak2012newton} and \cite{treister2014block}.
All these algorithms can be seamlessly applied here to solve the $\ell_1$-penalized $\bsbW$-optimization problem.
We use a  projected scaled sub-gradient algorithm due to \cite{schmidt2010graphical} (the Gafni-Bertsekas variant) which is developed on  the orthant-wise quasi-Newton method   \citep{andrew2007scalable}.

The complete procedure of estimating the association structure is presented in Algorithm \ref{Alg:W}. Note that the auxiliary matrix $\bsbC$ does not have to be explicitly computed at all, and forming the matrices $\bsbXi$ and $\bsbS$ does not need SVD or matrix square-root operations. This is because the key quantity $\bsbTh^{[k]}$ can be written as a (weighted) average of $\bsbM$ and $\bsbXi^{[k]}$\begin{align*}
\bsbTh^{[k]} &= \bsbM + \bsbC^{[k]}( \bsbI - \phi\bsbW^{[k]})^{1/2}\\ & = \bsbM + (\bsbXi^{[k]}-\bsbM)( \bsbI -\phi \bsbW^{[k]})^{1/2}( \bsbI -\phi \bsbW^{[k]})^{1/2}  \\& = \bsbM \bsbW^{[k]}\phi + \bsbXi^{[k]}(\bsbI - \phi\bsbW^{[k]}).
\end{align*}
  Moreover, with    $\bsbTh$,   $\bsbW$,  $\bsbXi$ available, evaluating the objective function value does not need    $\bsbC$, either.  This is because $Tr\{\bsbC\bsbW \bsbC^T\}=Tr\{(\bsbXi - \bsbM)  ( \bsbI -\phi \bsbW) \bsbW (\bsbXi - \bsbM)^T\}$. \

   When   $l(\theta, y)= (\theta-y)^2/2$, $\bsbXi^{[k]}$ in Step 4 is always fixed at $ \bsbY$, indicating  no need to iterate. Otherwise     GGL will be called iteratively. The key component of  Algorithm \ref{Alg:W} relies on  GGL that is well studied. Hence iGGL  provides great   implementation ease compared to designing a separate algorithm for each new problem with a different loss, and has  convergence guarantee according to Theorem \ref{lemma:lip}.

Experience shows that $\phi$ is not a sensitive parameter as long as it is small enough. (A further idea is   to use   varying  $\phi^{[k]}$,   say, $\phi^{[k]}=c /\| \bsbW^{[k]}\|_2$ with $c$ a small number less than 1. Although it shows   excellent performance in applications, we will investigate it in future work.)
When $\bsbM$ is unknown, e.g.,  $\bsbM=\bsb1\bsba^T$ with $m$ intercepts, the linearization still carries over and one can use BCD to solve for $\bsbM$ and $\bsbW$ alternatively in the $g$-optimization step.

\subsection{A Poisson re-parameterization} \label{ssec:poisson}

This subsection  concentrates on the Poisson case, i.e. $l(\theta,y)=-y\theta+\exp(\theta)$.
Since the loss does not have bounded curvature, there is a lack of universal stepsize.   In principle, this issue could be remedied
with line search but our experiments   show\  that it may  not be very  efficient.

We make an additional assumption that     $\boldsymbol{M}$ contains    intercepts, and write $\boldsymbol{M}=\boldsymbol{1}\boldsymbol{\alpha}^{T}+\boldsymbol{M}^{\circ}$.    This is a mild assumption in many applications, since      a Poisson random variable   cannot be centered  without changing its   distribution.  It follows that
\begin{equation}
\boldsymbol{\Theta}=\boldsymbol{1}\boldsymbol{\alpha}^{T}+\ensuremath{\bsbM^{\circ}+\bsbC( \bsbI- \phi\bsbW)^{1/2}}\equiv\boldsymbol{1}\boldsymbol{\alpha}^{T}+\bsbTh^{\circ},
\end{equation}
where $\boldsymbol{1}$ is a    column vector of $n$ ones and
$\boldsymbol{\alpha}\in\mathbb{R}^{m}$ represents the intercepts.
Plugging it into the loss,  we get
$
\bar{l}(\bsba, \bsbTh^\circ) = \sum_{k=1}^{m}   -\langle \bsby_k, \bsb1 \alpha_k +  \bsbth^\circ_k \rangle
+ \langle \bsb1, \exp(\bsb1 \alpha_k +   \bsbth^\circ_k) \rangle.
$
Let $\bsb{a} = \left[a_1, \cdots a_m\right]^T$ such that $
a_k = \alpha_k + \log \langle \bsb1, \exp(  \bsbth^\circ_k) \rangle
$, and
$\exp (a_k) = \langle \bsb1, \exp(\bsb1 \alpha_k +   \bsbth^\circ_k)\rangle$ with $\exp(\cdot)$ defined componentwise.
Then
\begin{align*}
\bar{l}(\bsba, \bsbTh^\circ) 
= & \sum_{k=1}^{m}
\left[-\langle \bsby_k,   \bsbth^\circ_k \rangle
+ \langle \bsby_k, \bsb1 \rangle \log \langle \bsb1, \exp(  \bsbth^\circ_k)\rangle \right]
+
\left[- \langle \bsby_k, \bsb1 \rangle a_k +\exp(a_k) \right],
\end{align*}
which is separable in \   $\bsb{a}$ and  $\bsbTh^\circ$. If there is no further penalty imposed on $\bsb{a}$, $
\bsb{a}^{opt} = \log(\bsbY^T \bsb1 ).
$
The loss on $\bsbTh^\circ$ writes
\begin{align*}
\bar{l}(\bsbTh^\circ) = \sum_{k=1}^{m} l_k(\bsbth_k^\circ) = \sum_{k=1}^{m}
-\langle \bsby_k,   \bsbth^\circ_k \rangle
+ c_k \log \langle \bsb1, \exp(  \bsbth^\circ_k)\rangle,
\end{align*}
where $c_k=\langle \bsby_k, \bsb1\rangle$.
 It is easy to see that the  gradient of $l_k$   with respect to $\bsbth^\circ_k$  is $
 -\bsby_k+ c_k \exp(\bsbth^\circ_k)/ \langle 1, \exp(\bsbth^\circ_k\rangle  ),
$
and so the associated Hessian   satisfies
\begin{equation*}
\mathcal H (\bsbth_k) = c_{k}\left\{ \textrm{diag}\left[\frac{\exp(\boldsymbol{\theta}_{k}^{\circ})}{\langle\boldsymbol{1},\exp(\boldsymbol{\theta}_{k}^{\circ})\rangle}\right]-\frac{\exp(\boldsymbol{\theta}_{k}^{\circ})[\exp(\boldsymbol{\theta}_{k}^{\circ})]^{T}}{\langle\boldsymbol{1},\exp(\boldsymbol{\theta}_{k}^{\circ})\rangle^{2}}\right\} \preceq \frac{c_k}{2} \bsbI .
\end{equation*}
 To secure the desirable properties associated with
$\rho=1$, one  can scale $l_k$   by  $c_k/2$ or use the overall Lipschitz constant      $L=\max_{1\le k \le m} \langle \bsby_k,\bsb1\rangle=\|\bsbY\|_1/2$. 

\section{Statistical  Analysis}
\label{sec:theory}

 In this section, we assume that $\bsbM$ is known and \eqref{lipcond} is satisfied with $L=1$. The overall objective in \eqref{crit:raw} is denoted by $F$ and the penalty $P$  takes the form of   $P(\bsbW) =  (n /2) \lambda \| \bsbW\|_1$.   Because of the nonconvex nature of the problem,  studying the performance of the set of global minimizers may not provide enough guidance in practice.   We will investigate the   statistical accuracy of the set of \textit{fixed points}  under the iGGL algorithm mapping (cf. \eqref{T+1} with $\rho=1$):
\begin{align}
\mathcal F=\Big\{(\hat\bsbW, \hat\bsbC)\in \mathcal Z: (\hat \bsbW,  \hat \bsbC) =   \argmin_{(\bsbW , \bsbC )\in \mathcal Z} g(\bsbW, \bsbC;  \bsbW^{- },  \bsbC^{- }) \big|_{\bsbW^- = \hat \bsbW, \bsbC^- = \hat \bsbC}\Big\}.\label{fixedpoint}
\end{align}
The $g$-minimization problem in \eqref{fixedpoint} has a unique minimizer.
In fact, given any feasible $\bsbW$, $g$ is strongly convex in $\bsbC$. Define $${\bsbC^\star}( \bsbW; \bsbW^-, \bsbC^-) = \argmin_{\bsbC} g(\bsbW, \bsbC; \bsbW^-, \bsbC^-).$$ Then   $g( \bsbW, \bsbC^\star( \bsbW; \bsbW^-, \bsbC^-); \bsbW^-,\bsbC^-)$ is strictly convex in $\bsbW$.

First,  we need to  define the    effective noise to take into account the randomness of observations.
Recall that the   marginal losses are ``arbitrarily'' chosen, and so we are not in  a standard   likelihood setting.
  Let $\bsbW^*  \in \mathcal{S}_{++}^m(\phi)$ denote the   statistical truth, which is the learning target. Then we can show that there  exists   $\bsbC^*$ satisfying $\bsbC^* = \bsbC^\star(\bsbW^*;   \bsbW^*, \bsbC^*)$; see,  Lemma \ref{existCstar}.
  Define
\begin{align*}
\bsbE = - \frac{2}{n}\frac{\rd (F-P) (  \bsbW, \bsbC^\star(\bsbW;  \bsbW^-, \bsbC^-))}{{\rd \bsbW}} \big|_{\bsbW = \bsbW^*,  \bsbW^-=\bsbW^*, \bsbC^- = \bsbC^*}
\end{align*}
 as the effective noise.
Hence in the noise-free scenario, the loss  as a function of  $\bsbW$   must vanish at the statistical truth.
 In the   Gaussian case,   $\bsbE$ becomes $   \bsbW^{* -1} - (\bsbY-\bsbM)^T(\bsbY - \bsbM)/n $.

   Next, we choose   a proper   discrepancy measure         to facilitate the  analysis.
 For two  matrices $\bsbW_1, \bsbW_2$ that  are both positive-definite, the  Bregman divergence  associated with $-\log \det(\cdot)$ is given by
\begin{align*}
\Breg (\bsbW_1,  \bsbW_2) = -\log\det (\bsbW_1) + \log\det (\bsbW_2) + \langle \bsbW_{2}^{-1}, \bsbW_1 - \bsbW_2\rangle.
\end{align*}
This divergence is always nonnegative since $-\log \det$\ is strictly convex. When $\bsbW_1, \bsbW_2$ are close, $\Breg (\bsbW_1,  \bsbW_2)\approx Tr\{  [(\bsbW_1 - \bsbW_2)\bsbW_2^{-1}]^2\}/2$ which resembles relative error.
We will use its symmetrized version  $${\Breg^{(s)}}(\bsbW_1, \bsbW_2) = (\Breg (\bsbW_1, \bsbW_2) + \Breg (\bsbW_2, \bsbW_1)  )/2 = \langle \bsbW_{2}^{-1} - \bsbW_{1}^{-1}, \bsbW_1 - \bsbW_2\rangle/2$$ to characterize  the error. In the following theorem, we use $\|\bsbE\|_{\max}$ to denote $\max_{j,k} |e_{j,k}|$. (As before, we assume   $\phi$ is chosen sufficiently small,  so that  $\|\bsbW^*\|_2 \le \phi^{-1}$ and $\|\hat\bsbW\|_2 \le \phi^{-1}$.) Define the support of $\bsbW^*$ by   $\mathcal J^*=\{(j, j'): w_{j,j'}^* \ne 0\}$ and    $J^* = |\mathcal J^*|$.
\begin{theorem}\label{th:errbnd} Assume there exist large enough  $K\ge 0, \vartheta>0$   such that
\begin{align}
(1+1/\vartheta) \| (\bsbW - \bsbW^*)_{\mathcal J^*}\|_1 \le \| (\bsbW - \bsbW^*)_{\mathcal J^{* c}}\|_1 +  K \{   J^* \Breg^{(s)} (\bsbW, \bsbW^*)\}^{1/2}\label{regcond}
\end{align}
 for any $\bsbW \in \mathcal S_{++}^m(\phi)$.
Then, on the event $\{\| \bsbE \|_{\max} \le \lambda_0\}$, with $\lambda = (2A +  2\vartheta+1) \lambda_0$  for any    $A\ge 0$, \emph{any} fixed point  $(\hat \bsbW, \hat\bsbC)\in \mathcal F$ satisfies
\begin{align}
  {\Breg^{(s)}(\hat \bsbW; \bsbW^*)} + Tr\{ (\hat \bsbW - \bsbW^*)(\bsbS(\hat \bsbW, \hat \bsbC) - \bsbS(\bsbW^*, \bsbC^*)\}   \le      (\vartheta+A)^2 K^2 \lambda_0^2 J^*, \label{errbnd}
\end{align}
where $\bsbS(\bsbW, \bsbC) = (\bsbI - \phi\bsbW)^{-1/2} \bsbC^T \bsbC (\bsbI - \phi\bsbW)^{-1/2}/n$.
 \end{theorem}

 \begin{corollary} \label{errbnd_subexp}
 Assume $e_{j,k}\sim \mbox{sub-exp}(c_1 /  n , c_2/n) $  for all $ 1\le j, k \le m$ (cf. Definition \ref{subexp}). Let  $\lambda =  c_1   (2A+ 2\vartheta+1)\{{(4+2\alpha) (\log m) /n}\}^{1/2}$ with   $A, \alpha\ge 0$.
  Then under \eqref{regcond} and  $n\ge ( c_2^2/c_1) (4+2\alpha) \log m$, \eqref{errbnd}  holds with probability $1-2m^{-\alpha}$ and  the error  bound on the right-hand side becomes $
(4+2\alpha)c_1^2    ( \vartheta+ A)^2  K^2   J^*(\log m) / n.
$
 \end{corollary}
 See Appendix \ref{apdx:stat} for the proof detail, from which the multiplicative constant  preceding the Bregman   term on the left-hand side of  \eqref{errbnd}  can be strengthened to any positive number less than $2$.

\eqref{regcond} is a comparison  regularity condition \citep{She2016} and  is implied by
\begin{align}
(1+1/\vartheta)^2  \| (\bsbW - \bsbW^*)_{\mathcal J^*}\|_1^2 \le   K^2  J^* \Breg^{(s)} (\bsbW, \bsbW^*) \label{comp}
\end{align}
or \begin{align}
(1+1/\vartheta)^2  \| (\bsbW - \bsbW^*)_{\mathcal J^*}\|_2^2 \le   K^2    \Breg^{(s)} (\bsbW, \bsbW^*) \label{RE}
\end{align}
for any $\bsbW \in \mathcal S_{++}^m(\phi): \| (\bsbW - \bsbW^*)_{\mathcal J^{* c}}\|_1< (1+1/\vartheta) \| (\bsbW - \bsbW^*)_{\mathcal J^*}\|_1  $.   The conditions of   \eqref{comp} and \eqref{RE} can be viewed as extensions of compatibility and  restricted-eigenvalue assumptions \citep{van2009conditions} in the graph setting. But our regularity condition is less restrictive.

In the Gaussian case,  the second   discrepancy term in trace form vanishes, and  so $\|\bsbW_1 - \bsbW_2 \|_F^2 \le \phi^2 Tr\{ \bsbW_{2}^{-1}  (\bsbW_1 - \bsbW_2) \bsbW_{1}^{-1} ( \bsbW_1 - \bsbW_2)\}=2\phi^2{\Breg^{(s)}}(\bsbW_1, \bsbW_2) ={\mathcal O}_p((\log m) \allowbreak J^*/ n) $, the rate of which  matches that of Theorem 1 in \cite{rothman2008sparse} if $K$, $A$, $\vartheta$, $\alpha$, $\phi$, $c_1, c_2$ are treated as constants.

\section{Experiments} \label{sec:exp}
\subsection{Synthetic data  }
Simulation experiments for non-Gaussian graph learning are performed in three settings: continuous variables with outlier contamination, binary variables, contaminated binary variables.
Given each setting, 50 i.i.d. datasets are generated and
we report the average results.
To evaluate the performance of each algorithm, we use the graph identification rate \citep{liu2012nonparanormal} which is defined as  the percentage of the correctly identified non-zero off-diagonal weights, i.e.,  $|\mathcal J_{\mbox{\tiny off}}(\hat \bsbW) \cap \mathcal J_{\mbox{\tiny off}}( \bsbW^*)|/|\mathcal J_{\mbox{\tiny off}}(\bsbW^*)|$, where $\mathcal J_{\mbox{\tiny off}}(\bsbW)$ denotes the indices of the nonzero off-diagonal entries of $\bsbW$.
To eliminate the interference  of various  tuning schemes and reveal the true potential
of each method, we vary the regularization parameter in some
pre-specified grids (100 grid values in all experiments) and use  the true cardinality to find the best estimate.
In calling iGGL, we used the $\ell_1$ penalty to enforce sparsity.  All experiments were performed on a machine with 2.1GHz
CPU and 16GB RAM.

The first setting is regarding continuous variables with outlier contamination.   Concretely, we first generated
a symmetric matrix with  standard Gaussian entries,  set
$90\%$ of its off-diagonal entries to zero, and then added  $ \eta\boldsymbol{I}$ to   make the    precision matrix $\bsbW^*$.    $\eta$ was chosen such that $\lambda_{\min}(\boldsymbol{W}^{*})=1$; see \cite{mazumder2012}. Next, we generated the observation matrix $\bsbY$ with each row following $\mathcal N(\bsb0,  \bsbW^{* -1} )$ and introduced gross outliers by modifying  $\alpha\%$ of the entries  to   $20$.
We set $n=1000$ and $m=10, 100$ (with the number of free parameters being   $55$ and $5050$, respectively). Table \ref{Gaussian simulation} shows a comparison between graphical lasso  and  iGGL with  Tukey's loss. The proposed method  was  much better at accommodating outliers in such  non-Gaussian graph learning and its stability  is impressive.

\begin{table}[H]
\begin{centering}
{\footnotesize 
\caption{\footnotesize   Graph identification rates ($\times 100$) for graphical lasso (g-lasso)
and  iGGL (Tukey) on outlier-contaminated Gaussian data, where $\alpha\%$ denotes the outlier percentage in the observation matrix. \label{Gaussian simulation}}

\begin{tabular}{c cccc c cccc}
\hline
 & \multicolumn{4}{c}{$m=10$} && \multicolumn{4}{c}{$m=100$}\tabularnewline
 \cline{2-5}  \cline{7-10}
$\alpha\%$ & $0\%$ & $5\%$ & $10\%$ & $20\%$ && $0\%$ & $5\%$ & $10\%$ & $20\%$\tabularnewline
\hline
g-lasso & $88$ & $18$ & $16$ & $16$ && $65$ & $15$ & $15$ & $15$\tabularnewline
iGGL & $88$ & $80$ & $76$ & $73$ && $65$ & $64$ & $64$ & $60$\tabularnewline
\hline
\end{tabular}
}\par\end{centering}
\end{table}

In the second setting,
 we  created binary data matrices according to the Ising
model  $p(y_{1}, y_{2},\ldots, y_{m}) \propto \exp(\sum_{j\neq i} y_{i} y_{j}w_{ij}^*+\sum_{i}b_{i}^* y_{i})$
and $y_{i}\in\{-1,+1\}$.
We set $b_{i}^*=0$ in common with most works, and generated     sparse $\{ w_{ij}^* \}$ ($50$ $w_{ij}^*$  chosen at random taking value $0.5$ and the remaining     $0$).
We compared    the  popular pseudo-likelihood based Ising graph learning  \citep{hofling2009estimation} with iGGL that applies Bernoulli deviance on each binary   variable.
Both methods \emph{approximate}   the genuine likelihood that is intractable. According to   Table \ref{Ising simulation_fixed},   pseudo-likelihood did an excellent job in this  setting, and iGGL, without making the Ising model assumption, showed strikingly similar  performance, which will be justified in a future paper.

\begin{table}[H]
\begin{centering}
{\footnotesize %
\caption{\footnotesize Graph identification rates ($\times 100$) for pseudo-likelihood based Ising
graph learning
and  iGGL (Bernoulli) on  binary data generated according to the  Ising model. \label{Ising simulation_fixed}}
\begin{tabular}{c ccc c ccc}
\hline
& \multicolumn{3}{c}{$n=100$} && \multicolumn{3}{c}{$n=200$}\tabularnewline
 \cline{2-4}  \cline{6-8}
$m$ & $100$ & $200$ & $400$ && $100$ & $200$ & $400$\tabularnewline
\hline
pseudo-likelihood & $87$ & $82$ & $74$ && $99$ & $98$ & $98$\tabularnewline
iGGL & $87$ & $83$ & $72$ && $99$ & $98$ & $98$\tabularnewline
\hline
\end{tabular}
}\par\end{centering}

\end{table}

Furthermore, we studied   non-Gaussian graph learning in  a contaminated binary  setting. The  binary observations
were generated according to the aforementioned scheme, but we  modified $\alpha\%$ of the rows of the data matrix to $[1,\ldots 1]$.
    Table \ref{Bernoulli simulation outlier} shows the graph identification results of pseudo-likelihood and iGGL that uses the robust Lorenz loss. The flexibility and universality  of iGGL offered significant performance improvement in this challenging setup.

\begin{table}[H]
\begin{centering}
{\footnotesize
\caption{\footnotesize Graph identification rates ($\times 100$) for pseudo-likelihood based Ising
graph learning
and  iGGL (Lorenz) on  contaminated binary data, where     the   percentage of outlier rows is denoted by  $\alpha\%$. \label{Bernoulli simulation outlier}}
\begin{tabular}{c ccc c ccc}
\hline
& \multicolumn{3}{c}{$m=50$} && \multicolumn{3}{c}{$m=200$}\tabularnewline
 \cline{2-4}  \cline{6-8}
$\alpha\%$ & $5\%$ & $10\%$ & $20\%$ &&  $5\%$ & $10\%$ & $20\%$\tabularnewline
\hline
pseudo-likelihood   & $83$ & $75$ & $49$ &&   $77$ & $67$ & $35$\tabularnewline
iGGL  & $86$ & $81$ & $63$ &&   $80$ & $71$ & $44$\tabularnewline
\hline
\end{tabular}
}
\par\end{centering}

\end{table}

Finally, we   conducted experiments to investigate the scalability of the iGGL algorithm by varying    problem dimensions. Here,  we set   $n=5000$ and    $m=100, 200, 500, 1000, 2000$ (the number of free parameters being  $5.05\mbox{e+}3$,  $2.01\mbox{e+}4$, $1.25\mbox{e+}5$, $5.01\mbox{e+}5$, $2.01\mbox{e+}6$, respectively).   Using a  grid of 100 points  for $\lambda$, we computed the solution path till  the model cardinality  increases to $n/(\log m^{2})$ (as suggested   by the proof of Theorem \ref{th:errbnd}).
Table \ref{Computational efficiency without Acc} shows the computational time (in seconds) averaged over $50$ independent
simulations,
when   we tested Tukey's loss, Bernoulli deviance, and reparametrized Poisson loss in the iGGL framework. Nicely, the computational time
increases  {linearly}  with the number of parameters, an evidence of the scalability of iGGL for non-Gaussian graph learning.


\begin{table}[H]
\begin{centering}
{\footnotesize
\caption{\footnotesize Computational time (in CPU seconds) when applying Tukey's loss, Bernoulli deviance, and reparametrized
Poisson loss (denoted by Tukey, Binary, Discrete, respectively).}
\label{Computational efficiency without Acc}
\begin{tabular}{cccccc}
\hline
$m$ & $100$ & $200$ & $500$ & $1000$ & $2000$\tabularnewline
\hline
Tukey    & $7.9$ & $16.9$ & $50.4$ & $131.7$ & $535.8$\tabularnewline
Binary   & $5.2$ & $10.7$ & $32.6$ & $85.3$ & $338.7$\tabularnewline
Discrete & $5.1$ & $11.9$ & $41.8$ & $122.2$ & $511.5$\tabularnewline
\hline
\end{tabular}
}
\par\end{centering}
\end{table}


\subsection{S\&P 500 }
This dataset keeps a record of the closing prices of S\&P 500 stocks
from Jan. 1, 2003 to Jan. 1, 2008 \citep{Zhao2012_huge}. It consists
of $1258$ samples for  $452$  stocks and has been
preprocessed by taking logarithm and differencing transformations.

\begin{figure}[H]
\centering{}\includegraphics[width=1\columnwidth,height=1\textheight,keepaspectratio]{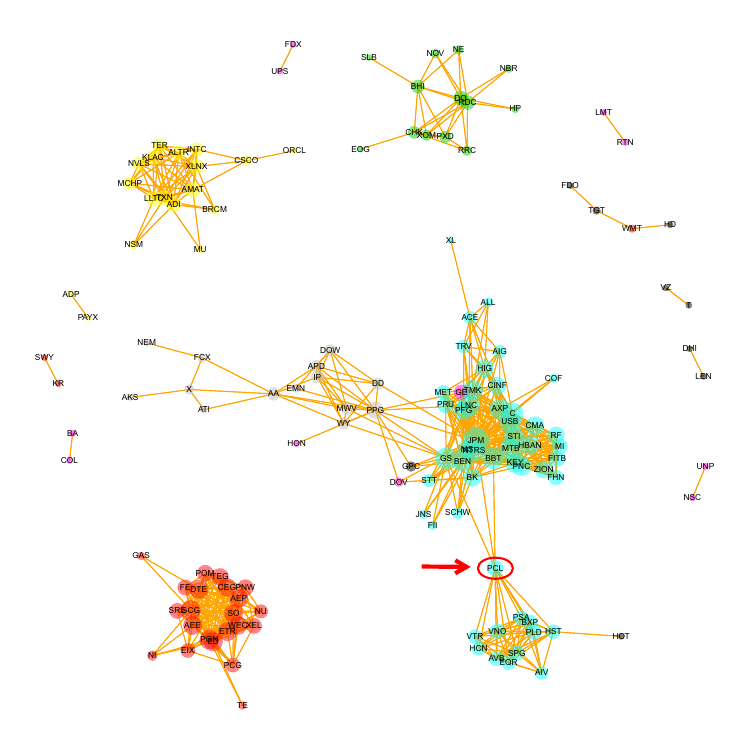}\caption{Robust graph learning by iGGL on S\&P 500. \label{PIC Tukey} }
\end{figure}

We   chose     Tukey's robust loss in learning the association
graph   by iGGL due to some potential outliers occurring in such financial data.
(It is well known that even for the transformed data, anomalies, caused
by extreme market movements, may be present.)   PIC   \citep{She2017selective}  was used for parameter tuning, with the degrees-of-freedom   $\mbox{df} = |\mathcal J_{\mbox{\tiny off}}(\hat \bsbW)|$ and the inflation given by $\mbox{df}\cdot\log (em (m-1)/\mbox{df})$.  The overall computational time was 3 minutes. Figure \ref{PIC Tukey}
demonstrates the topological structure after removing all isolated
nodes. The nodes were colored    according to the Global Industry
Classification Standard (GICS) sectors provided in the dataset documentation.
Overall, the  clusters revealed in Figure \ref{PIC Tukey}
 are pretty consistent with the GICS sectors  in different colors. For example, at the top,   \texttt{INTC}, \texttt{ADI},
\texttt{TXN} and other  semiconductor companies  (in yellow) form a subgraph, so are  oil related
companies (in green) including \texttt{XOM}, \texttt{DO} and \texttt{BHI}, and at the bottom is a big cluster of utilities  companies (in red) such as \texttt{WEC}, \texttt{SCG} and \texttt{AEP}.

Notably, a number of financial companies, \texttt{JPM}, \texttt{BBT},
\texttt{AXP}, among many others   near the
center of the graph are densely connected. An interesting fact is that  they connect to the
estate companies  in the lower-right corner through
\texttt{PCL}, i.e., Plum Creek Timber, which was
the largest private owner of timberland in the U.S..
Also,     \texttt{GE}, though belonging  to the sector of industrials (in magenta),  is found   in the financial cluster. The documentation indicates that the       GE still included  the large financial division GE Capital  during the
data collection period.

\begin{figure}[H]
\begin{centering}
\subfloat[g-lasso with $nnz = 500, 1000, 2000$]{\begin{centering}
\includegraphics[width=0.35\columnwidth,height=0.175\textheight,keepaspectratio]{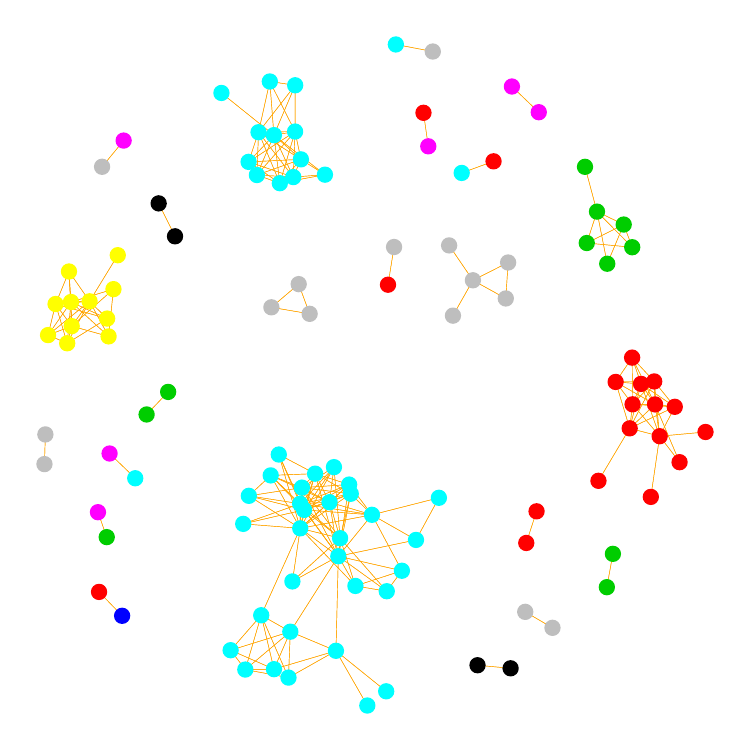}\includegraphics[width=0.35\columnwidth,height=0.175\textheight,keepaspectratio]{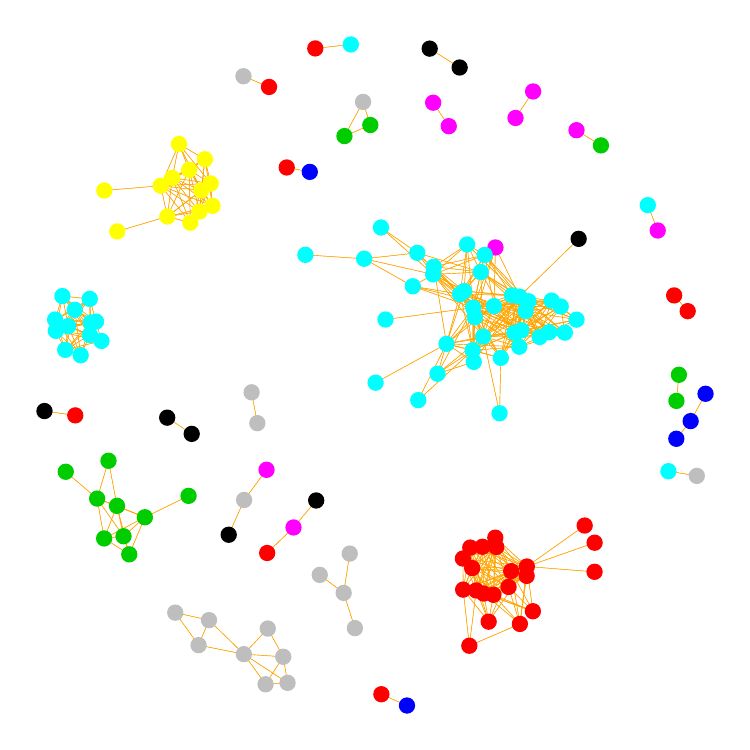}\includegraphics[width=0.35\columnwidth,height=0.175\textheight,keepaspectratio]{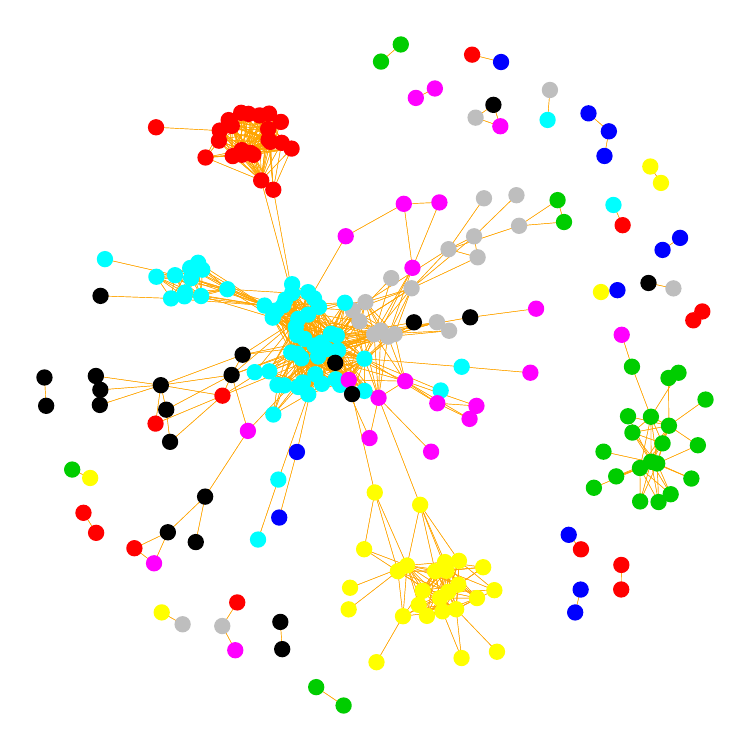}
\par\end{centering}
}
\par\end{centering}
\begin{centering}
\subfloat[iGGL with $nnz = 500, 1000, 2000$]{\begin{centering}
\includegraphics[width=0.35\columnwidth,height=0.175\textheight,keepaspectratio]{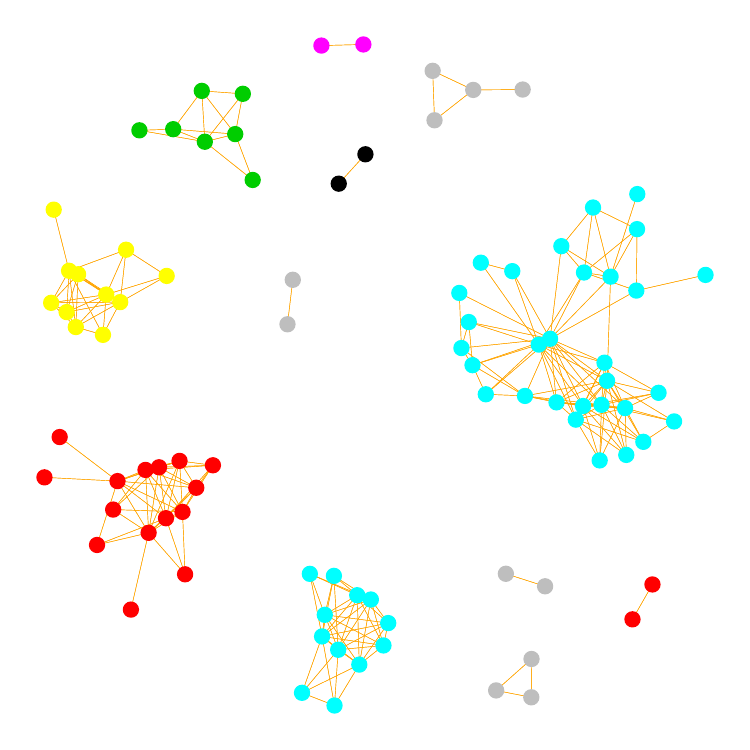}\includegraphics[width=0.35\columnwidth,height=0.175\textheight,keepaspectratio]{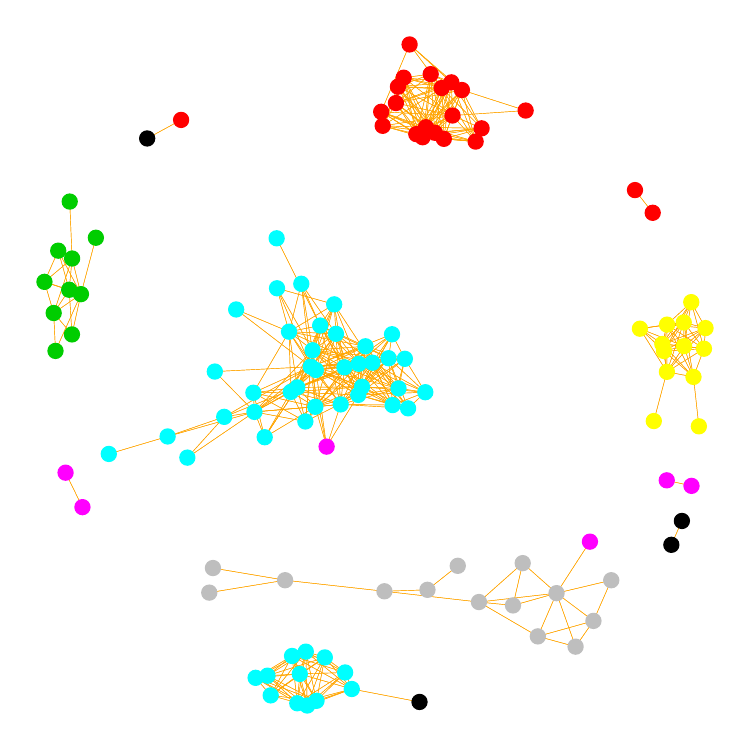}\includegraphics[width=0.35\columnwidth,height=0.175\textheight,keepaspectratio]{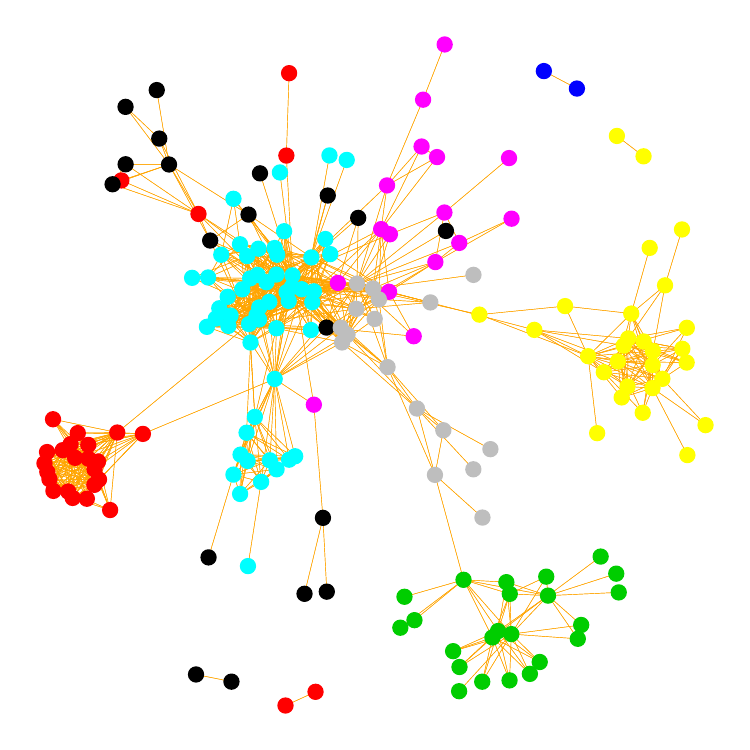}
\par\end{centering}
}
\par\end{centering}
\begin{centering}
\subfloat[non-paranormal g-lasso with $nnz = 500, 1000, 2000$]{\begin{centering}
\includegraphics[width=0.35\columnwidth,height=0.175\textheight,keepaspectratio]{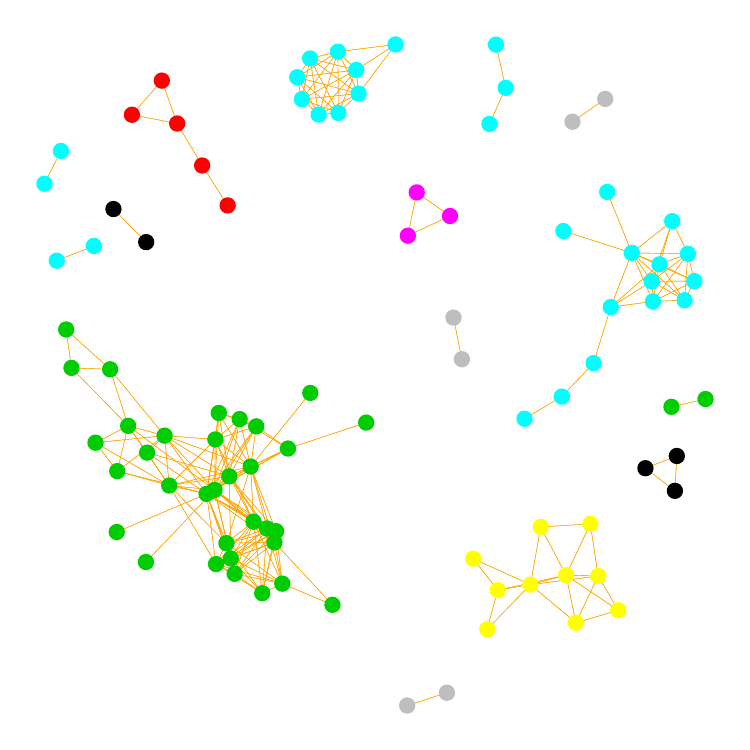}\includegraphics[width=0.35\columnwidth,height=0.175\textheight,keepaspectratio]{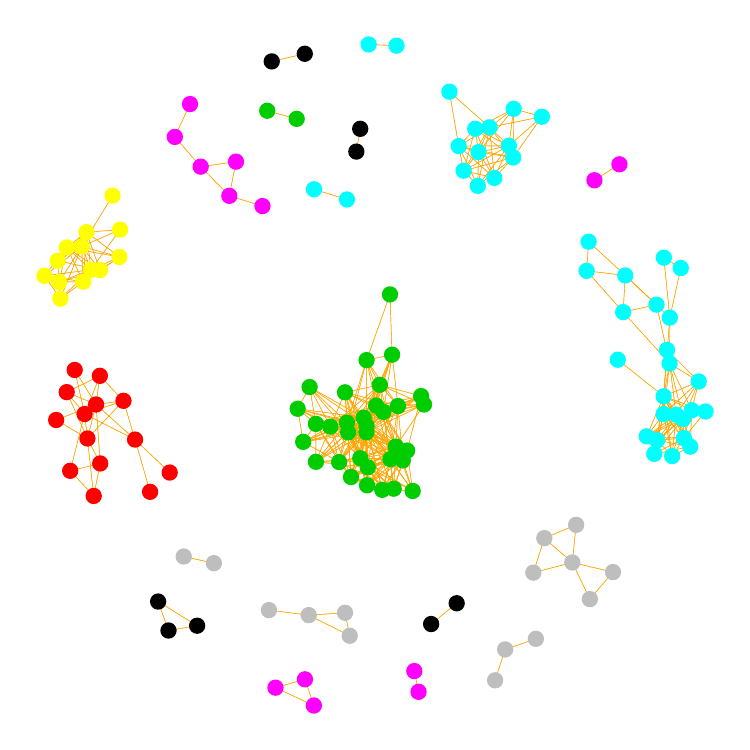}\includegraphics[width=0.35\columnwidth,height=0.175\textheight,keepaspectratio]{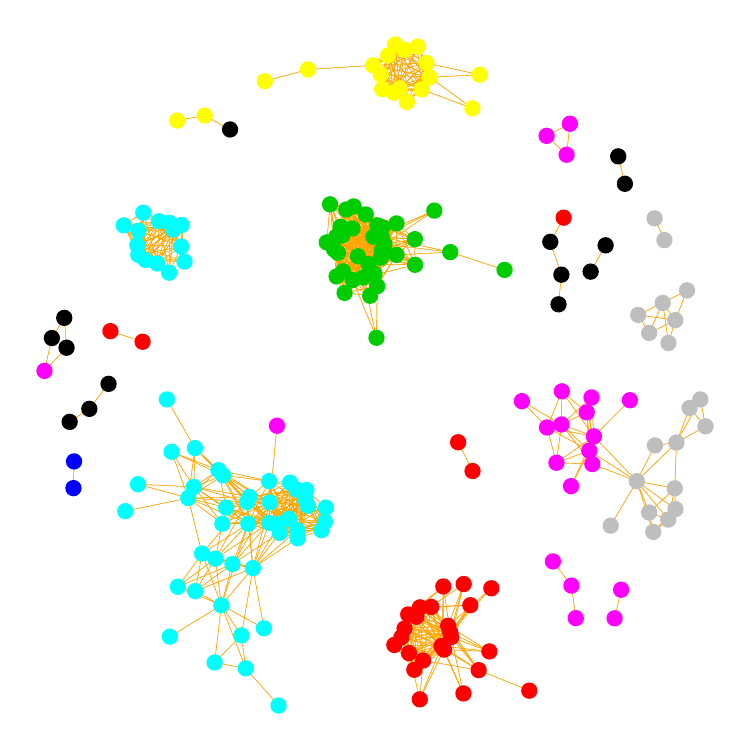}
\par\end{centering}
}
\par\end{centering}
\begin{centering}
\subfloat[non-paranormal neighborhood with $nnz = 500, 1000, 2000$]{\begin{centering}
\includegraphics[width=0.35\columnwidth,height=0.175\textheight,keepaspectratio]{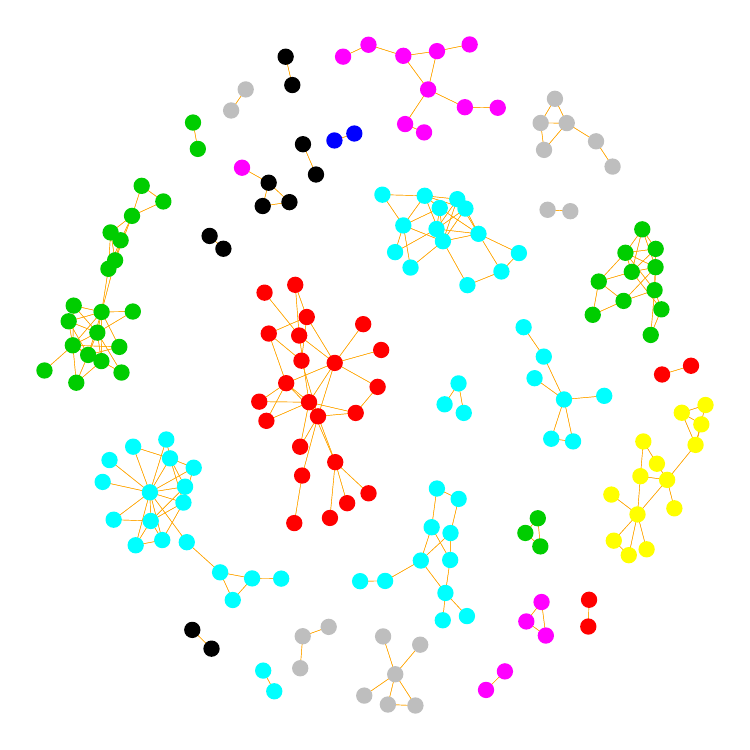}\includegraphics[width=0.35\columnwidth,height=0.175\textheight,keepaspectratio]{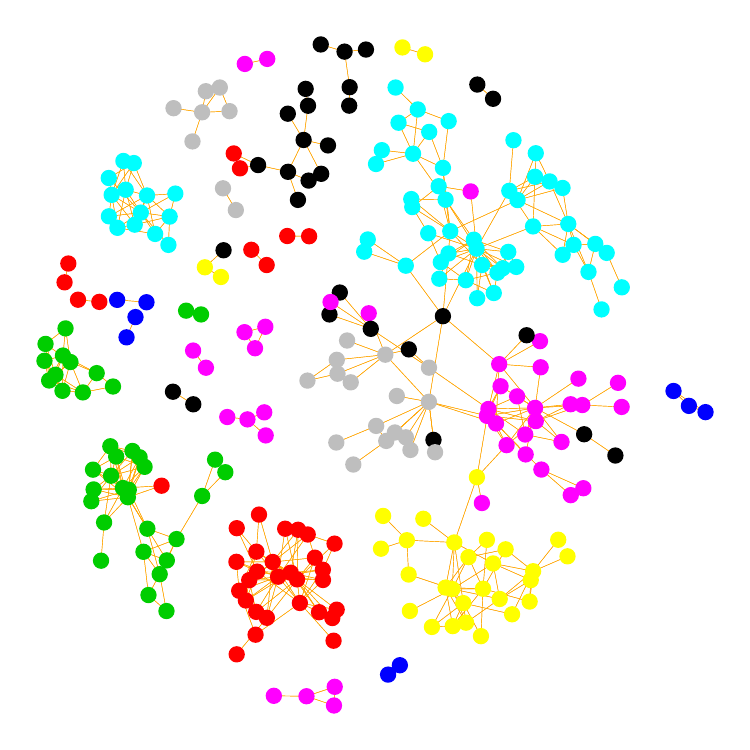}\includegraphics[width=0.35\columnwidth,height=0.175\textheight,keepaspectratio]{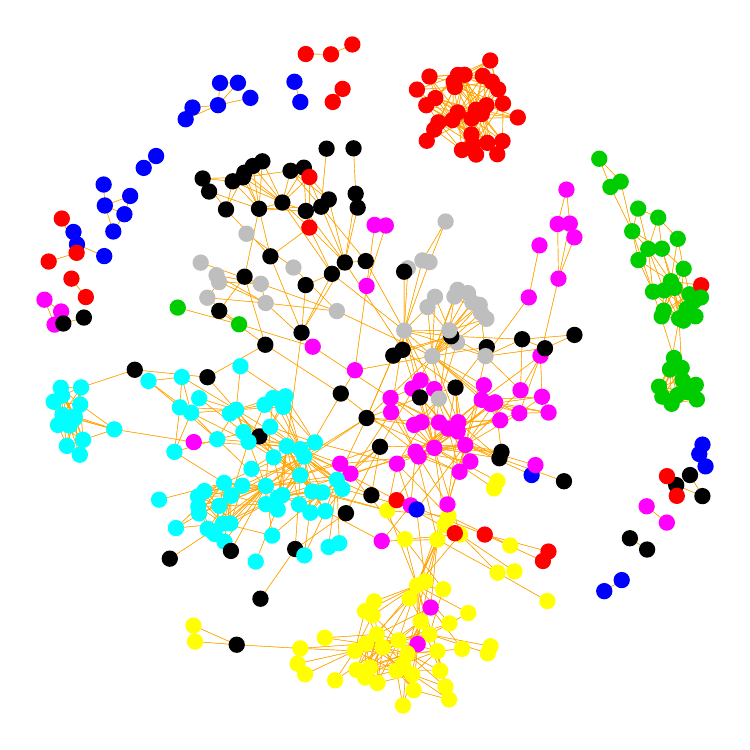}
\par\end{centering}
}
\par\end{centering}
\centering{}\caption{ Graphical lasso, iGGL, non-paranormal
graphical lasso, and non-paranormal neighborhood pursuit, with     cardinality ($nnz$, number of nonzero off-diagonal entries) aligned at   $500$, $1000$ and
$2000$, from left to right. All isolated nodes have been removed.\label{Solution path comparison}}
\end{figure}

We also experimented with  graphical lasso, non-paranormal graphical lasso, and non-paranormal  neighborhood pursuit on S\&P 500. Some graph estimates   along the solution path are demonstrated in   Figure \ref{Solution path comparison}. 
When the  cardinality is   small (say   $500$), seen from the colors, the obtained subgraph    structures  comply with the SICS sectors to a large extent. But the  graphical lasso estimates are less `pure'   in color.
Between the non-paranormal neighborhood and non-paranormal graphical lasso, many empirical studies in the literature (for example,  \cite{Tuo2014})  favor the first. We found that  when the cardinality reaches $2000$,   the first algorithm exhibits richer inter-section connections;   iGGL-Tukey   shared the same feature as shown in   Figure \ref{Solution path comparison}.
We also performed $100$ bootstrap experiments  to compare iGGL with nonparanormal graphical lasso. 
 Using a cutoff frequency of 60\%,    we found the nonparanormal    rank-based algorithm    shows no direct or indirect connection between \texttt{HIG} (Hartford Financial Services Group) and  \texttt{JPM} (JPMorgan Chase), while the two  are  directly connected in the  iGGL graphs  96\% of the time.
Then, we  performed non-parametric tests based on Kendall's tau and Spearman's rho:  both suggest significant  associations  between   \texttt{JPM} and \texttt{HIG}, with   $p$-values smaller than $1\textrm{e-}15$.

\subsection{TDT2}
\begin{figure}[h!]
 \begin{center}
 \includegraphics[width=1\columnwidth]{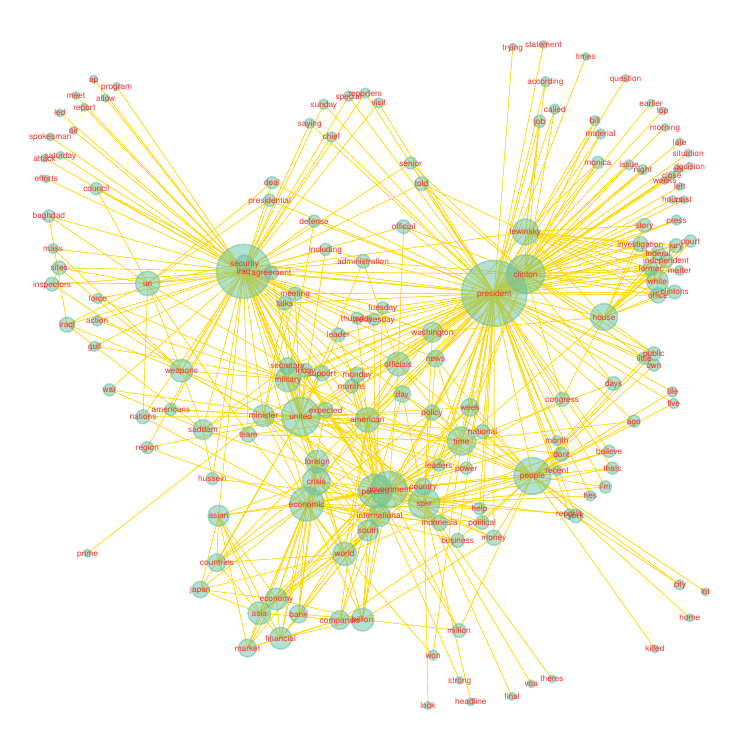}
 \caption{ TDT2 graph with no isolators. \label{fig:tdt2part}}
 \end{center}
 \end{figure}

The data is  provided by  \cite{cai2005document}. It is a subset of the   TDT2 corpus  collected  during the first half of  1998 from   6 sources:    2 newswires ({\textsc{APW}}, {\textsc{NYT}}), 2 radio programs ({\textsc{VOA}}, {\textsc{PRI}}) and 2 television programs ({\textsc{CNN}}, {\textsc{ABC}}).
We picked  200  most frequently used words, and the
document-term matrix records   frequencies ({counts})  of each term in $n=9,394$  documents.
 We ran   iGGL on the count data and used PIC for parameter  tuning. The computation of the solution path and tuning took   about 19  minutes. A sparse association graph is plotted in 
Figure \ref{fig:tdt2part} where  only the connected nodes  are shown for a better view.


As seen in Figure \ref{fig:tdt2part},  \texttt{president} makes a big hub and   connects to \texttt{congress}, \texttt{policy}, \texttt{officials}, \texttt{washington}, and so on. Right next to it is    \texttt{clinton} which also  exhibits rich connections to a variety of terms. In particular, it has a direct association with   \texttt{lewinsky}  which   is linked to  \texttt{monica}, \texttt{story}, \texttt{told}, etc.

On the left, we see many words  around  \texttt{iraq} and  \texttt{un}, and these   two words    share  some common neighbors like   \texttt{weapons}  and  \texttt{deal}.    In addition, interesting connections   exist between \texttt{crisis}, \texttt{asia},  \texttt{financial}, and \texttt{market} in the lower part of the graph.
These  reflect  some   hot topics and key news events  in   the first half of 1998.

\subsection{Newsgroup data}

\begin{figure}[h!]
\centering{}\includegraphics[width=1.2\columnwidth]{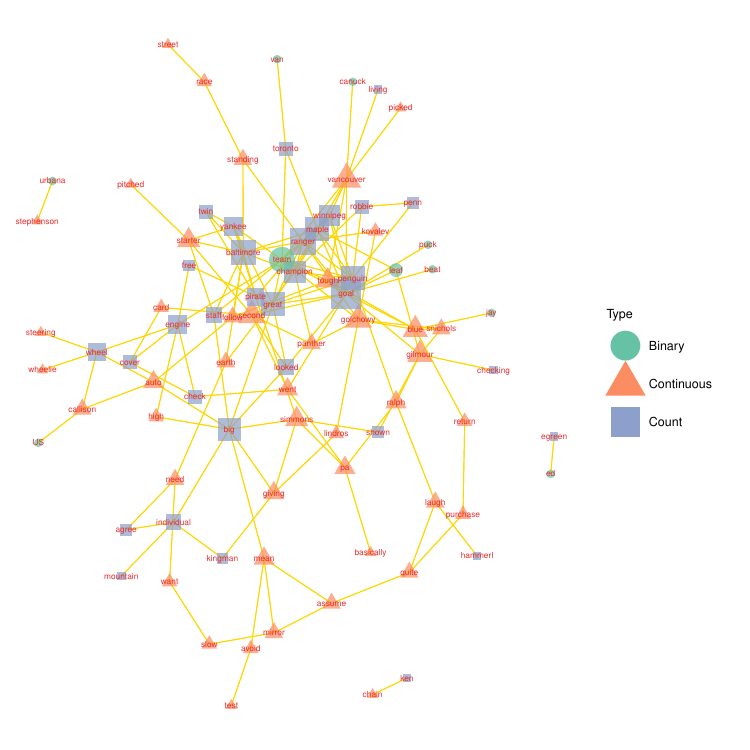}
\caption{Mixed graph learning on newsgroups. \label{fig:newsgroup-graph}}
\end{figure}

We use the recreation  subset of   the $20$ newsgroups data including the newsgroups of        \emph{rec.autos},
\emph{rec.motorcycles}, \emph{rec.sport.baseball}, and   \emph{rec.sport.hockey},\footnote{The 20 newsgroup dataset is available at \texttt{http://qwone.com/{$\sim$}jason/20Newsgroups/}.}
  which  gives   $2,389$ documents. After the pre-processing
(tokenization) with   {Python} packages  {scikit}-{learn}
and {NLTK} \citep{Bird2009}, we obtained  $36,365$
words in total, many of which, however,    seem to  have little   importance or meaning. We performed a word
filtering based on the       TF-IDF (term
frequency-inverse document frequency) statistics  which are widely used         in text mining. Furthermore, the words with non-alphabetical letters were   removed and
we    picked $150$ words as the variables of interest. These words were divided  into three
equal-sized subsets  for mixed graph learning:   the words in the first subset   take
TF-IDF values as the observed data, with    Tukey as the loss type;   the second subset   uses       word counts;   for the third subset, only    the information of word occurrence
  (binary)   is   kept, and   Lorenz is applied  as the marginal losses. We notice that the mixed data types posed a more challenging problem in computation---it took us about 2.4 hours to complete the solution path computation and parameter tuning.   Figure \ref{fig:newsgroup-graph}
shows the topological structure after removing  all isolated points.

As seen in the graph,       \texttt{team}   locates near the center which  indeed shows up in many   central topics  in the threads of      \emph{rec.sport.baseball}
 and  \emph{rec.sport.hockey}. To the right it has connections to
\texttt{winnipeg},  \texttt{maple}, \texttt{ranger},
\texttt{penguin},  
  relating to some famous
teams in the {{National Hockey League (NHL), while }}its left neighbors    \texttt{twin},
\texttt{yankee}, \texttt{baltimore}, and \texttt{pirate}     refer to  some  {Major League Baseball} (MLB) teams.
The graph also demonstrates   interesting connections between the words    that  mostly appear in       \emph{rec.autos}
and \emph{rec.motorcycles}. For example,
 \texttt{wheelie}, \texttt{steering}, \texttt{engine}, \texttt{auto},
\texttt{callison}    on the left side of the figure    are  linked   with  the keyword \texttt{wheel}.  The word \texttt{callison}  represents   \textit{James P. Callison}, who appears to be an expert in performance tuning and is a big fan of car racing  based on the conversations.    At the bottom,   \texttt{slow}, \texttt{mirror},
\texttt{avoid}  reflect a major concern in auto  and motorcycle
sports discussions---driving safety.

 \section{Conclusion}
\label{sec:conc}
In this work, we described the indirect Gaussian graph learning  framework  that is applicable to non-Gaussian data and has rich applications. An easy-to-implement optimization algorithm was developed based on iterative Gaussian graph learning. Our statistic algorithm analysis provides     provable guarantees for the associated solutions. The technique of over-parametrization with shrinkage is a universal and powerful tool to extend marginal estimation   to multivariate modeling without the need of  specifying a joint distribution. A range of modern machine learning problems  can be addressed  by jointly estimating the mean and dependency matrices in the proposed  framework which will be   investigated further.


\appendix

\section{Proof of Theorem \ref{lemma:equiv}}

        Let $f_1$ be the objective function in problem \eqref{f1}.   We can evaluate   the optimal  ${\bsbC}$ by letting $\nabla_{\bsbC} f_1 = 0 $:
        \begin{align*}
        -(\bsbY - \bsbM - \bsbC( \bsbI- \phi \bsbW)^{1/2})( \bsbI- \phi \bsbW)^{1/2} + \phi (\bsbC  \bsbW+\bsbC \bsbW^T)/2 = 0.
        \end{align*}
        Hence
        \begin{align*}
        \hat{\bsbC} =  \argmin_{\bsbC}  f_1
        =  (\bsbY-\bsbM)( \bsbI-\phi \bsbW)^{1/2}.
        \end{align*}
        Plugging $\hat{\bsbC}$ in  $f_1$, we have
        \begin{align*}
        f_1
        = & \frac{1}{2 \phi } Tr\{(\bsbY-\bsbM)(\bsbI -\bsbI+\phi \bsbW ))
        (\bsbI -\bsbI+\phi \bsbW )^T(\bsbY-\bsbM)^{T}\}\\
        & +\frac{1}{2 } Tr\{(\bsbY  - \bsbM)( \bsbI-\phi \bsbW)^{1/2}\bsbW (( \bsbI- \phi \bsbW)^{1/2})^T(\bsbY  - \bsbM)^T \}\\
        &-\frac{n}{2}\log \det \bsbW + P_M(\bsbM; \lambda_M) + P_W(\bsbW ; \lambda_W)\\
        = &\frac{ \phi }{2  } Tr\{(\bsbY  - \bsbM)\bsbW \bsbW(\bsbY  - \bsbM)^T\}\\&-\frac{1}{2   } Tr\{(\bsbY  - \bsbM)( \bsbW-\phi \bsbW\bsbW)^{1/2}( \bsbW-\phi \bsbW\bsbW)^{1/2}(\bsbY  - \bsbM)^T \}\\
        &-\frac{n}{2}\log \det \bsbW + P_M(\bsbM; \lambda_M) + P_W(\bsbW ; \lambda_W)\\
        = &\frac{1}{2   } Tr\{(\bsbY  - \bsbM) \bsbW(\bsbY  - \bsbM)^T\} - \frac{n}{2}\log \det \bsbW + P_M(\bsbM; \lambda_M) + P_W(\bsbW ; \lambda_W),
        \end{align*}
        which is exactly the objective function  in \eqref{f2}.
The conclusion follows if we set $\phi$ to be no more than $ 1/ \| \hat \bsbW \|_2 $ for any $\hat { \bsbW} $ as a solution to $
\min_{ \bsbM\in \mathcal X,    \bsbW   \in \mathcal{S}_{++}^m}    Tr\{(\bsbY  - \bsbM) \bsbW  (\bsbY  - \bsbM)^T\}/2 - (n/2)\log \det ( \bsbW ) + P_M(\bsbM; \lambda_M) + P_W( \bsbW; \lambda_W)$.


\section{Proof of Theorem \ref{lemma:asym}}
\label{apdx:asym}
Let $F$ be the objective function in the theorem. It is not difficult to calculate its gradients with respect to $\bsbC$ and $\bsbW$ (details omitted):
\begin{align*}
\nabla_{\bsbC} F  = & \phi^{-1} \nabla \bar l(\bsbTh) (\bsbI - \phi \bsbW)^{1/2} + \bsbC \bsbW \\
\nabla_{\bsbW} F  = & - (\bsbI - \phi \bsbW)^{-1/4} \frac {\bsbC^T \nabla \bar l(\bsbTh) +   \nabla \bar l(\bsbTh)^T \bsbC}{4}(\bsbI - \phi \bsbW)^{-1/4} + \frac{\bsbC^T\bsbC}{2} - \frac{n}{2}\bsbW^{-1}. 
 \end{align*}
Let $\bsbZ = \bsbI - \phi \bsbW $. Then the  optimal $\bsbC$ must satisfy $\nabla \bar l (  \bsbTh) = - \phi\bsbC  \bsbW \bsbZ^{-1/2}$, which, when plugged  into $\nabla_{\bsbW} F = \bsb{0}$, yields
$$
\frac{1}{4} [\bsbZ^{-1/4} \bsbC^T \bsbC (\phi \bsbW)\bsbZ^{-3/4} + \bsbZ^{-3/4} (\phi \bsbW)\bsbC^T \bsbC \bsbZ^{-1/4}]+ \frac{\bsbC^T\bsbC}{2} - \frac{n}{2}\bsbW^{-1} = \bsb{0}.
$$  Let $\bsbXi - \bsbM = \bsbC \bsbZ^{-1/2}$. Then
\begin{align}
& \frac{1}{4} [\bsbZ^{1/4} (\bsbXi - \bsbM)^T (\bsbXi - \bsbM)  \phi \bsbW \bsbZ^{-1/4} + \bsbZ^{-1/4}  \phi \bsbW (\bsbXi - \bsbM)^T (\bsbXi - \bsbM)\bsbZ^{1/4}] \notag \\
&+ \frac{1}{2}  \bsbZ^{1/2} (\bsbXi - \bsbM)^T (\bsbXi - \bsbM) \bsbZ^{1/2} - \frac{n}{2}\bsbW^{-1} = \bsb{0}. \label{weq_mod1}
\end{align}
Noticing that \begin{align}(\bsbI - \phi \bsbW)^{1/2} \bsbW^{-1} (\bsbI - \phi \bsbW)^{1/2}=\bsbW^{-1} (\bsbI - \phi \bsbW) = \bsbW^{-1} - \phi \bsbI,\label{comm1}\end{align} it  is easy to verify that   $\bsbW = [(\bsbXi - \bsbM)^T (\bsbXi - \bsbM)/n]^{-1}$ satisfies \eqref{weq_mod1}.
Hence in the following, we study the equation group
\begin{align}
\begin{cases}
\phi^{-1} \nabla \bar l(\bsbTh) \bsbZ^{1/2} + \bsbC \bsbW = \bsb{0}\\
\bsbW^{-1} =   (\bsbXi - \bsbM)^T (\bsbXi - \bsbM)/n\\
\bsbTh = \bsbM + \bsbC \bsbZ^{1/2}\\
\bsbXi - \bsbM = \bsbC \bsbZ^{-1/2}\\
\bsbZ = \bsbI -\phi\bsbW.
\end{cases}\label{gradeq_mod}
\end{align}

Recall the assumption that  $\bsbM = \bsb1 \bsba^T$ with $\bsba = [\alpha_1, \ldots, \alpha_m]^T$ and $l_k(\bsbth_k) = l_k(\theta_{1, k}) + \cdots + l_k(\theta_{n, k})  $.   By definition,
\begin{align*}
\nabla \bar l(\bsbTh) =[\nabla l_1(\bsbth_1), \ldots, \nabla l_m(\bsbth_m)]= \nabla \bar l (\bsbM) + \bsbC \bsbZ^{1/2}  \bsbD + \breg,
\end{align*}
where   $\breg$ is short for $\breg(\bsbTh; \bsbM)$.
The 1st equation in \eqref{gradeq_mod} becomes
\begin{align*}
( \nabla \bar l (\bsbM) + \bsbC \bsbZ^{1/2}  \bsbD + \breg) \bsbZ^{1/2} + \phi \bsbC \bsbW = \bsb0.
\end{align*}
Therefore,
\begin{align*}
\bsbC \bsbZ^{-1/2}& =  -  (\nabla \bar l (\bsbM)+\breg)  \bsbZ^{1/2} ( \bsbZ^{1/2} \bsbD \bsbZ^{1/2}  +   \phi  \bsbW)^{-1} \bsbZ^{-1/2} \\
& =-  (\nabla \bar l (\bsbM)+\breg) [\bsbZ\bsbD + \bsbZ^{1/2} \phi \bsbW \bsbZ^{-1/2}]^{-1}\\& =-  (\nabla \bar l (\bsbM)+\breg) [(\bsbI - \phi \bsbW)\bsbD + \phi \bsbW  ]^{-1}\\& =-  (\nabla \bar l (\bsbM)+\breg) [\bsbD + \phi \bsbW (\bsbI - \bsbD)]^{-1},
\end{align*}
where the third equality, similar to \eqref{comm1}, can be verified by   spectral decomposition.   The conclusion   follows from the 2nd and the 4th equations in \eqref{gradeq_mod}.

 \section{Proof of Theorem \ref{lemma:lip}}
\label{apdx:lip}
The proof is straightforward. For completeness,  some  details are given as follows.
We denote the   objective function and the surrogate function by $f(\bsbTh)$ and $g(\bsbTh;\bsbTh^{[k]})$, respectively, with a bit abuse of notation.
From the construction of the surrogate function,  we have
\begin{align}
g(\bsbTh^{[k+1]}; \bsbTh^{[k]}) \leq g(\bsbTh^{[k]}; \bsbTh^{[k]}) = f(\bsbTh^{[k]}).\label{nonincrease1}
\end{align}
It suffices to show
$f(\bsbTh) \leq g(\bsbTh; \widetilde\bsbTh)$ for any $\bsbTh, \widetilde\bsbTh$ and $\rho \geq L$,
or
        \begin{align}
        \bar{l}(\bsbTh) -
        \bar{l}(\widetilde\bsbTh) -
        \langle \nabla \bar{l}(\widetilde\bsbTh), \bsbTh - \widetilde\bsbTh\rangle
        -  \frac{\rho}{2}\|\bsbTh - \widetilde\bsbTh\|_F^2\leq 0.
        \end{align}

        It follows from the Lipschitz condition that
        \begin{align*}
        &\bar{l}(\bsbTh) - \bar{l}(\widetilde\bsbTh)- \langle \nabla \bar{l}(\widetilde\bsbTh), \bsbTh - \widetilde\bsbTh \rangle \\
        = &  \int_{0}^{1}
        \langle \nabla \bar{l}(\widetilde\bsbTh+t(\bsbTh - \widetilde\bsbTh)) ,
        \bsbTh - \widetilde\bsbTh \rangle \rd t
        - \int_{0}^{1}\langle \nabla \bar{l}(\widetilde\bsbTh), \bsbTh - \widetilde\bsbTh \rangle \rd t\\
        = & \int_{0}^{1} \langle \nabla \bar{l}(\widetilde\bsbTh+t(\bsbTh - \widetilde\bsbTh))
        - \nabla \bar{l}(\widetilde\bsbTh), \bsbTh - \widetilde\bsbTh \rangle \rd t \\
        \leq & \int_{0}^{1} \sum_{i,j} L t |\theta_{ij} - \tilde\theta_{ij}|^2 \rd t  \\
        \leq & \frac{L}{2} \|\bsbTh - \widetilde\bsbTh\|_F^2.
        \end{align*}
The conclusion thus follows.

\section{Proofs  of Theorem \ref{th:errbnd} and Corollary  \ref{errbnd_subexp}}
\label{apdx:stat}

\begin{lemma} \label{basicfacts} The following basic facts hold
\begin{enumerate}
\item  $g( \bsbW, \bsbC; \bsbW, \bsbC) = F( \bsbW, \bsbC)$;
\item
 $\frac{\partial g( \bsbW^+, \bsbC^+;  \bsbW, \bsbC)}{\partial \bsbC^+}|_{ \bsbW^+ =\bsbW, \bsbC^+ = \bsbC} = \frac{\partial F(\bsbW, \bsbC)}{\partial \bsbC}$;
 \item  $\frac{\partial (g-P)( \bsbW^+, \bsbC^+;  \bsbW, \bsbC)}{\partial \bsbW^+}|_{ \bsbW^+ =\bsbW, \bsbC^+ = \bsbC} = \frac{\partial (F-P)( \bsbW, \bsbC)}{\partial \bsbW}$; \item $ \frac{\partial g( \bsbW^+, \bsbC^+;  \bsbW, \bsbC}{\partial \bsbC^+} |_{\bsbC^+ = \bsbC^\star(\bsbW^+;  \bsbW, \bsbC)}= \bsb{0}$;
\item $\frac{\rd (g-P)( \bsbW, \bsbC^\star(\bsbW; \bsbC^-,\bsbW^-);   \bsbW^-, \bsbC^-)}{\rd \bsbW}    = (n/2) ( \bsbS( \bsbW, \bsbC^\star(\bsbW; \bsbC^-, \bsbW^-)) - \bsbW^{-1} )$, where $\bsbS(\bsbW, \bsbC) = (\bsbI - \phi \bsbW)^{-1/2} \bsbC^T \bsbC (\bsbI - \phi \bsbW)^{-1/2}/n$.
\end{enumerate}
\end{lemma}

These properties  are easy to verify from    the construction of $g$, and the proofs are  omitted. 

Recall that ${\bsbC^\star}( \bsbW; \bsbW^-, \bsbC^-) = \argmin_{\bsbC} g(\bsbW, \bsbC; \bsbW^-, \bsbC^-)$, where we omit its dependence on data matrix $\bsbY$. To guarantee that $\bsbE$ is well defined we need the following lemma.
\begin{lemma}\label{existCstar}
Given any feasible $\bsbW^*\in   \mathcal{S}_{++}^m$, there exists $\bsbC^*$  such that $\bsbC^* = \bsbC^\star(\bsbW^*;   \bsbW^*, \bsbC^*)$.
\end{lemma}

To prove the result, define a sequence of iterates $$\bsbC^{k+1} =\bsbC^\star(\bsbW^*;   \bsbW^*, \bsbC^k)=
\min_{\bsbC} g(\bsbW^*, \bsbC; \bsbW^*, \bsbC^{k}).$$ Let  $f(\bsbC) = F(\bsbW^*, \bsbC)$ and $h(\bsbC, \bsbC^-) = g(\bsbW^*, \bsbC; \bsbW^*, \bsbC^-)$.  Define ${\breg}_{\bar l}(\bsbC, \bsbC')  = \bar l(\bsbC) - \bar l(\bsbC') - \langle \nabla \bar l(\bsbC'), \bsbC - \bsbC'\rangle$ and $\Breg_2(\bsbC, \bsbC')= \|\bsbC - \bsbC'\|_F^2/2$. Then it is easy to see that
\begin{align}
f(\bsbC^{k+1} )+ (\frac{1}{\phi} \Breg_2 - \frac{1}{\phi} \breg_{{\bar l}})(\bsbTh^{k+1}, \bsbTh^{k}) = h(\bsbC^{k+1}, \bsbC^k).\label{Cseqres1}
\end{align}
Moreover, from the optimality of $\bsbC^{k+1}$, we have
$$
h(\bsbC^{k+1}, \bsbC^{k}) + \frac{1}{2 \phi} \|\bsbC^k - \bsbC^{k+1}\|_{(\bsbI - \phi\bsbW^*)^{1/2}}^2 +  \frac{1}{2  } \|\bsbC^k - \bsbC^{k+1}\|_{ (\bsbW ^{*})^{1/2}}^2  \le h(\bsbC^k, \bsbC^k)
$$
or
\begin{align}
h(\bsbC^{k+1}, \bsbC^{k}) + \frac{1}{2 \phi} \|\bsbC^k - \bsbC^{k+1}\|_F^2  \le  f(\bsbC^k).\label{Cseqres2}
\end{align}
Combining \eqref{Cseqres1} and \eqref{Cseqres2}  gives
\begin{align*}
\frac{1}{\phi} (2 \Breg_2 - \breg_{{\bar l}})(\bsbTh^k, \bsbTh^{k+1}) \le  f(\bsbC^{k}  ) - f(\bsbC^{k+1} ).
\end{align*}
It follows from the Lipschitz condition that
\begin{align}
 \|  \bsbTh^k -  \bsbTh^{k+1}\|_F^2   \le 2\phi(f(\bsbC^{k}  ) - f(\bsbC^{k+1} ))
\end{align}
and so the sequence of $f(\bsbC^k)$ is monotonically non-increasing. This implies that (i) $\bsbC^{k+1} - \bsbC^{k}\rightarrow \bsb0$, and (ii) $Tr\{\bsbC^k \bsbW^* \bsbC^k\}$ and thus $\bsbC^k$ are uniformly bounded. Therefore, any limit point of $\{\bsbC^{k}\}$ can serve as $\bsbC^*$ which also depends on $\bsbY$.
\\

From Lemma \ref{basicfacts}, we get
\begin{align*}
&\frac{\rd (F-P)(\bsbW, \bsbC^\star(\bsbW; \bsbW^-, \bsbC^-)) }{\rd \bsbW}\bigg|_{\bsbW=\bsbW^*, \bsbW^- = \bsbW^*, \bsbC^-=\bsbC^* }\\
= &    \frac{\partial F(\bsbW, \bsbC)}{\partial \bsbC} \big|_{\bsbW = \bsbW^*, \bsbC=\bsbC^*}   \frac{\rd \bsbC^\star(\bsbW; \bsbC^-, \bsbW^-)}{\rd \bsbW}\big|_{\bsbW = \bsbW^*, \bsbW^-=\bsbW^*, \bsbC^- = \bsbC^*} \\& +\frac{\partial (F-P)(\bsbW, \bsbC)}{\partial \bsbW}\big|_{\bsbW = \bsbW^*, \bsbC=\bsbC^*}. \end{align*}
For the first term,
\begin{align*}
\frac{\partial F(\bsbW, \bsbC)}{\partial \bsbC}\big|_{\bsbW = \bsbW^*, \bsbC=\bsbC^*}&\overset{(ii)}  = \frac{\partial g(\bsbW^+, \bsbC^+; \bsbW, \bsbC)}{\partial \bsbC^+}\big|_{\bsbW^+ =\bsbW^*, \bsbC^+=\bsbC^*, \bsbW= \bsbW^*, \bsbC = \bsbC^*} \\&\overset{(iv)} = \bsb0 .
\end{align*}
For the second term,
\begin{align*}
&\frac{\partial (F-P)(\bsbW, \bsbC)}{\partial \bsbW}\big|_{\bsbW = \bsbW^*, \bsbC=\bsbC^*} \\  \overset{(iii)}= & \frac{\partial (g-P)(\bsbW^+, \bsbC^+; \bsbW, \bsbC)}{\partial \bsbW^+}  \big|_{\bsbW^+=\bsbW^*, \bsbC^+ = \bsbC^*, \bsbW= \bsbW^*, \bsbC=\bsbC^*}\\
= & \frac{\rd (g-P)(\bsbW, \bsbC^\star(\bsbW; \bsbW^-, \bsbC^-); \bsbW^-, \bsbC^-)}{ \rd \bsbW}\big|_{\bsbW = \bsbW^*, \bsbC^-=\bsbC^*, \bsbW^- =\bsbW^*}\\ \overset{(v)} = & (n/2)(\bsbS(\bsbW^*, \bsbC^*) - \bsbW^{* -1}),
\end{align*}
where the second equality is due to
\begin{align*}
&\frac{\rd (g-P)(\bsbW, \bsbC^\star(\bsbW; \bsbW^-, \bsbC^-); \bsbW^-, \bsbC^-)}{\rd \bsbW} \\
=& \frac{\partial (g-P)(\bsbW, \bsbC; \bsbW^-, \bsbC^-)}{\partial \bsbC}\big|_{\bsbC= \bsbC^\star(\bsbW; \bsbW^-, \bsbC^-)} \frac{\rd \bsbC^\star(\bsbW; \bsbW^-, \bsbC^-)}{\rd \bsbW}\\&+ \frac{\partial (g-P)(\bsbW, \bsbC; \bsbW^-, \bsbC^-)}{\partial \bsbW}\big|_{\bsbC = \bsbC^\star(\bsbW; \bsbW^-, \bsbC^-)}\\ \overset{(iv)}= &  \frac{\partial (g-P)(\bsbW, \bsbC; \bsbW^-, \bsbC^-)}{\partial \bsbW}\big|_{\bsbC = \bsbC^\star(\bsbW; \bsbW^-, \bsbC^-)}.
\end{align*}
Therefore, we have
$\bsbE =    \bsbW^{* -1}-\bsbS(\bsbW^*, \bsbC^*)  $.

 Given any fixed point $(\hat \bsbW, \hat\bsbC)$ under the algorithm mapping, we have $g(\hat \bsbW, \hat\bsbC; \hat \bsbW , \hat \bsbC ) \le  g(  \bsbW,  \bsbC; \hat \bsbW , \hat \bsbC )$, $\forall (\bsbW,\bsbC)\in \mathcal Z$. Taking $\bsbC = \hat \bsbC$, we know from Theorem \ref{lemma:equiv} that   $ \hat \bsbW $ is the global minimizer of the following problem
\begin{align*}
 \min_{\bsbW\in \mathcal S_{++}^m}
&
\frac{n}{2 }
Tr\{ \bsbS(\hat \bsbW, \hat \bsbC)
\bsbW  \}
-\frac{n}{2}\log \det (\bsbW) +  P(\bsbW; \lambda).
\end{align*}
For short  write  $\hat \bsbS$  for  $\bsbS(\hat \bsbW, \hat \bsbC)$.
Since the loss and the penalty are  convex, it is not difficult to show
\begin{align*}
&  \frac{n}{2} Tr\{\hat \bsbS \hat\bsbW\} - \frac{n}{2} \log \det (\hat\bsbW) + P(\hat\bsbW)+ \frac{n}{2} \Breg(\bsbW; \hat\bsbW) \\ \le \ &  \frac{n}{2} Tr\{\hat\bsbS \bsbW\} - \frac{n}{2} \log \det (\bsbW)+P(\bsbW),
\end{align*}
where $\Breg$ is the Bregman divergence associated with $-\log\det$.  Then   \begin{align*}
    \frac{n}{2}  Tr\{\hat \bsbS (\hat\bsbW-\bsbW)\} + \frac{n}{2} \Breg (\hat\bsbW, \bsbW)  + \frac{n}{2} \Breg(\bsbW, \hat\bsbW) \\+\frac{n}{2} \langle - \bsbW^{-1}, \hat\bsbW - \bsbW\rangle   \le    P(\bsbW) -  P(\hat\bsbW),
\end{align*} and setting $\bsbW = \bsbW^*$ yields
\begin{align*}
   n {\Breg^{(s)}} (\hat\bsbW, \bsbW^*) +  \frac{n}{2} Tr\{\hat \bsbS (\hat\bsbW-\bsbW^*)\}  + \frac{n}{2} \langle - \bsbW^{*-1}, \hat\bsbW - \bsbW^*\rangle \\ \le       P(\bsbW^*) -  P(\hat\bsbW).
\end{align*}
Denote   $  \bsbS(\bsbW^*, \bsbC^*)$ by $\bsbS^* $. Then
\begin{align}
&    n {\Breg^{(s)}} (\hat\bsbW; \bsbW^*) + \frac{n}{2} Tr\{(\hat \bsbS  - \bsbS^*)(\hat\bsbW-\bsbW^*)\}   \notag\\
\le \,&   P(\bsbW^*) -  P(\hat\bsbW)   + \frac{n}{2} \langle   \bsbW^{*-1} - \bsbS^*, \hat\bsbW - \bsbW^*\rangle \notag\\
 =\, &    P(\bsbW^*) -  P(\hat\bsbW)   + \frac{n}{2} \langle \bsbE , \hat\bsbW - \bsbW^*\rangle.\label{l1pens}
\end{align}
 Under $ \| \bsbE  \|_{\max} \le \lambda_0 $, \eqref{l1pens} is bounded by $    ({n}/{2})( \lambda \| \bsbW^*\|_1 - \lambda \| \hat\bsbW\|_1 + \lambda_0 \| \hat \bsbW - \bsbW^*\|_1)$.
It follows from   the definition of $\mathcal J^*$ and the sub-additivity of the $\ell_1$-norm that
\begin{align*}
& 2 {\Breg^{(s)}} (\hat\bsbW; \bsbW^*) +  Tr\{(\hat \bsbS  - \bsbS^*)(\hat\bsbW-\bsbW^{*})\}   \\ \le \, &    ( \lambda + \lambda_0)  \| (\bsbW^* - \hat\bsbW)_{\mathcal J^*}\|_1 - (\lambda - \lambda_0) \| (\hat \bsbW - \bsbW^*)_{\mathcal J^{*c}}\|_1 \\
= \, &   \lambda_0 ( 2 A + 2 \vartheta +  2)   \| (\bsbW^* - \hat\bsbW)_{\mathcal J^*}\|_1 -  \lambda_0 (2 A + 2\vartheta )   \| (\hat \bsbW - \bsbW^*)_{\mathcal J^{*c}}\|_1,
\end{align*}
where we set $\lambda =  (2A + 1+2 \vartheta) \lambda_0$ with $A\ge 0$. Using the regularity condition and Cauchy-Schwarz inequality, we have
\begin{align*}
& 2 {\Breg^{(s)}} (\hat\bsbW; \bsbW^*) +  Tr\{(\hat \bsbS  - \bsbS^*)(\hat\bsbW-\bsbW)\}  \\ \le & \lambda_0 2(  A +  \vartheta )    K (J^* \Breg^{(s)}(\hat\bsbW, \bsbW^*))^{1/2}\\ \le  & \frac{   2(\vartheta+A)^{2}   K^2 \lambda_0 ^2 J^*}{  a} + \frac{a}{2}    \Breg^{(s)}(\hat\bsbW, \bsbW^*)
\end{align*}
for any $a>0$. Taking $a=2$ gives the desired result.\\

To prove     Corollary  \ref{errbnd_subexp}, we first define sub-exponential random variables:
\begin{definition}\label{subexp}
$X$ is   sub-exponential with mean $0$ and parameters $(\nu, b)$, i.e.,  $ X\sim \mbox{sub-exp}(\nu, b)$,  if and only if there exist $\nu, b\ge 0$ such that $\EE [\exp(\lambda  X )] \le \exp(\nu \lambda^2/2)$ for all $|\lambda|\le 1/b$.
\end{definition}
A basic property of $X\sim \mbox{sub-exp}(\nu, b)$ is  that $\EP[|X|\ge t]\le 2 \exp(-t^2/(2\nu))$ for any $t: 0\le t\le \nu/b$ which is easy to show based on the definition.

Therefore, under $e_{j,k} \sim \mbox{sub-exp}(c_1/  n , c_2/n) $, $\forall j, k: 1\le j, k \le m$
\begin{align*}
\EP(\max |e_{j,k}  | \ge t) \le &  2\exp(-\frac{ n t^2}{2c_1} + 2\log m ), \forall t : 0\le t \le \frac{c_1}{c_2}\end{align*}
Let  $t = \lambda_0 = A_0 \sqrt{(\log m)/n}$ with $A_0:  A_0^2 = c_1(4+2\alpha)$. Then the above probability bound is $2m^{-\alpha}$ as long as  $A_0 \sqrt{(\log m )/ n }\le c_1/ c_2$ or $n\ge    (4+2\alpha)(c_2^2/c_1)\log m $.

\bibliographystyle{apalike}
\bibliography{IGGL}

\end{document}